\documentclass[runningheads]{llncs}
\usepackage{graphicx}

\usepackage{tikz}
\usepackage{comment}
\usepackage{amsmath,amssymb} 
\usepackage{color}
\usepackage{orcidlink}

\usepackage[accsupp]{axessibility}  

\usepackage{subfig}
\usepackage{booktabs}
\usepackage{adjustbox}
\usepackage{multirow}
\usepackage{wrapfig}

\newcommand{\argmax}[1]{ \underset{#1}{\operatorname{argmax}}~ }

\newcommand{\ie}[0]{\textit{i.e.}}
\newcommand{\eg}[0]{\textit{e.g.}}
\newcommand{\etal}[0]{\textit{et al}}
\newcommand*\samethanks[1][\value{footnote}]{\footnotemark[#1]}

\begin{document}
\pagestyle{headings}
\mainmatter
\def\ECCVSubNumber{2512}  

\title{Constructing Balance from Imbalance for Long-tailed Image Recognition} 

\titlerunning{Constructing Balance from Imbalance}
%
\author{
Yue Xu\inst{1}\orcidlink{0000-0001-7489-7269}\thanks{The first two authors contribute equally.} 
\and
Yong-Lu Li\inst{1,2}\orcidlink{0000-0003-0478-0692}\samethanks
\and
Jiefeng Li\inst{1}\orcidlink{0000-0003-1932-8914} 
\and
Cewu Lu\inst{1}\orcidlink{0000-0003-1533-8576}\thanks{Cewu Lu is the corresponding author, member of Qing Yuan Research Institute and Shanghai Qi Zhi institute.}
}
\authorrunning{Y. Xu et al.}
%
\institute{Shanghai Jiao Tong University
\and
Hong Kong University of Science and Technology\\
\email{\{silicxuyue,yonglu\_li,ljf\_likit,lucewu\}@sjtu.edu.cn}
}

\maketitle

\begin{abstract}
    Long-tailed image recognition presents massive challenges to deep learning systems since the imbalance between majority (head) classes and minority (tail) classes severely skews the data-driven deep neural networks. Previous methods tackle with data imbalance from the viewpoints of data distribution, feature space, and model design, etc.
    In this work, instead of directly learning a recognition model, we suggest confronting the bottleneck of head-to-tail bias before classifier learning, from the previously omitted perspective of \textbf{balancing label space}.
    To alleviate the head-to-tail bias, we propose a concise paradigm by progressively adjusting label space and dividing the head classes and tail classes, \textit{dynamically constructing balance from imbalance} to facilitate the classification. 
    With flexible data filtering and label space mapping, we can easily \textbf{embed} our approach to most classification models, especially the decoupled training methods. 
    Besides, we find the separability of head-tail classes varies among different features with different inductive biases. Hence, our proposed model also provides a \textit{feature evaluation} method and paves the way for long-tailed \textbf{feature} learning. 
    Extensive experiments show that our method can boost the performance of state-of-the-arts of different types on widely-used benchmarks.
    \textbf{Code is available at https://github.com/silicx/DLSA}.
\end{abstract}


\section{Introduction}
\label{sec:intro}

\begin{figure}[!t]
	\centering
    \subfloat[Linear Classifier]{\includegraphics[width=0.3\linewidth]{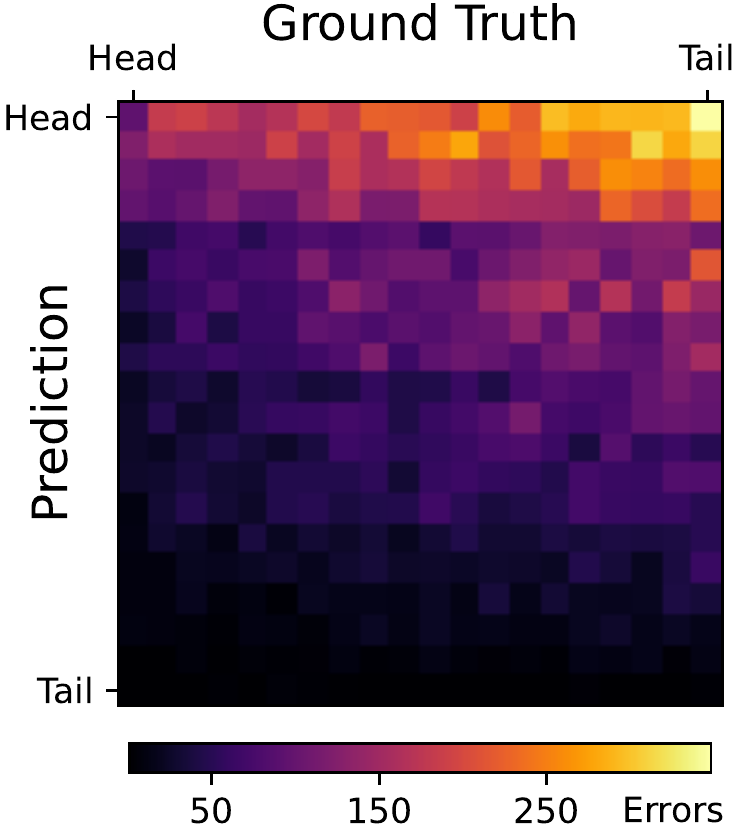}}
    \subfloat[PaCo~\cite{paco}]{\includegraphics[width=0.3\linewidth]{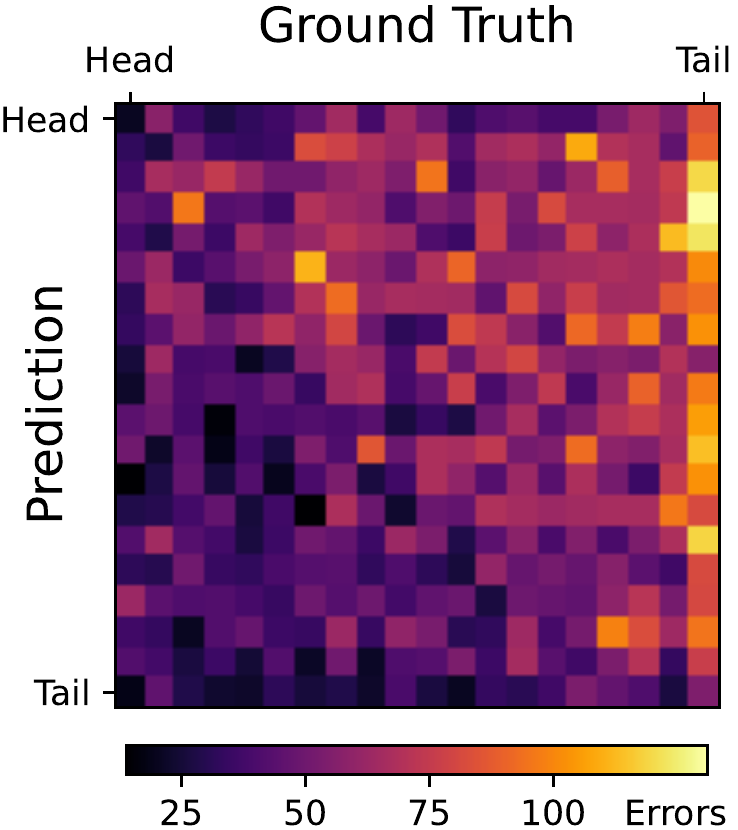}}
	\caption{Confusion matrices of models on ImageNet-LT~\cite{openlongtail} test set, indicating the severe head-tail bias and that tail classes are particularly prone to confusion with head classes (high density at right-top). \textbf{The correct samples are omitted for clarity}. The classes are ordered by their frequency and merged into 20 bins.
	}
	\label{fig:error-confusion}
\end{figure}

Deep learning shows its superiority in various computer vision tasks~\cite{vgg,alexnet,resnet}, especially in balanced data scenarios.
Though, real-world data is usually severely imbalanced, following a long-tailed distribution~\cite{zipf2013psycho,spain2007measuring,li2022hake,li2020pastanet}, \ie, very few frequent classes take up the majority of data (head) while most classes are infrequent (tail).
The highly biased data skews classifier learning and leads to performance drop on tail classes.
As shown in Fig.~\ref{fig:error-confusion}, most errors stem from the \textit{head-to-tail bias} and a large number of tail samples are misclassified as head classes, even with the very recent long-tail learning technique~\cite{paco}.

Many approaches have been proposed to re-balance long-tail learning by balancing the data distribution~\cite{japkowicz2002class,he2009learning,buda2018systematic,byrd2019effect,Remix,MetaSAug}, balancing the output logits~\cite{menon2020long,zhong2021improving}, balancing the training losses~\cite{huang2016learning,huang2019deep,wang2017learning,LDAM,BALMS,samuel2021distributional,sinha2020class}, balancing the feature space \cite{yang2020rethinking,kang2020exploring,wang2021contrastive,jiang2021self,paco} or with balanced training strategy~\cite{decouple,zhou2020bbn,jamal2020rethinking}.
However, as it is the definition of class labels to blame for the long-tailed data distribution, there are very few works tackling long-tailed recognition from the perspective of balancing the \textit{label space}. 
Samuel~\etal~\cite{samuel2021generalized-attribute} decompose the class labels into semantic class descriptors and exploit its familiarity effect, which is commonly used in zero-shot learning. Wu~\etal~\cite{wu2020solving-taxonomy} reorganize the class labels into a tree hierarchy by realistic taxonomy.
Both methods partly alleviate the label imbalance but are restricted by the \textit{class setting} (semantics, hierarchy). Here, we want to dynamically adjust the label space according to the imbalance of \textit{realistic data distribution} to fit long-tail learning. 

\begin{figure*}[ht]
\begin{minipage}{0.40\textwidth}
	\centering
	\includegraphics[width=\textwidth]{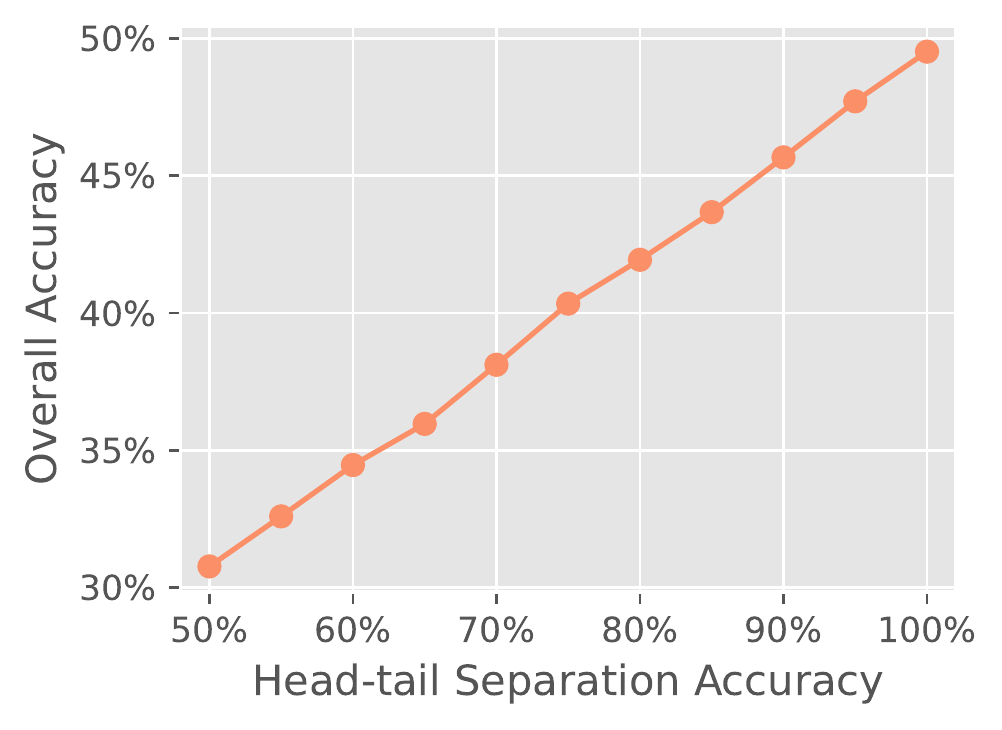}
	\caption{The performance curve of linear classifier on ImageNet-LT~\cite{openlongtail} with different head-tail separation accuracy (50\%: \textbf{random} separation; 100\%: \textbf{ideal} separation). The separation model divides the classes into head and tail, and then we classify within the two individual groups. \textbf{Better} separation leads to a \textbf{higher} overall accuracy.
	}
	\label{fig:separate-class}
\end{minipage}
\hfill
\begin{minipage}{0.57\textwidth}
    \centering
	\includegraphics[width=\textwidth]{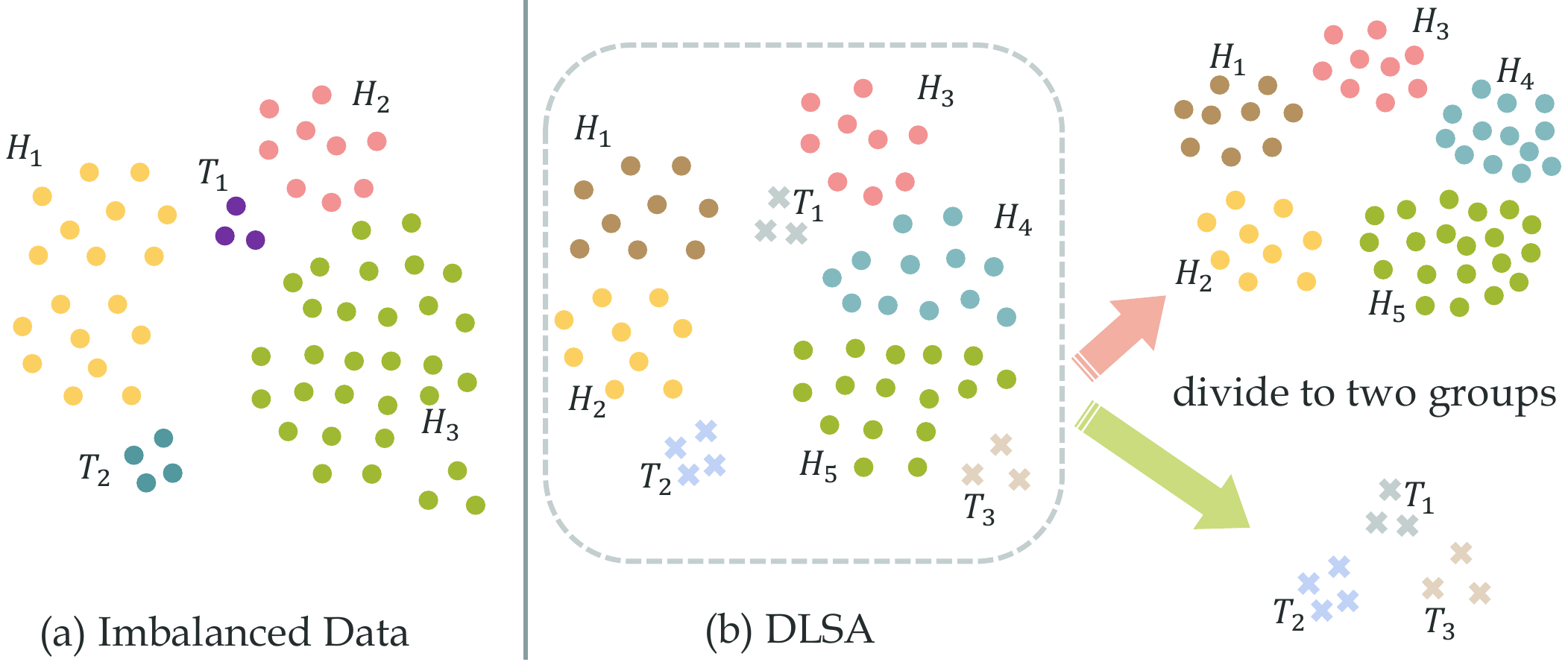}
	\caption{Demonstration of our dynamic label space adjustment (DLSA). $H_i$ stands for a head class and $T_j$ is a tail class. We re-define label space for imbalanced data and divide head and tail classes into two groups progressively to reduce head-to-tail bias, seeking balance from the new data sub-space. DLSA not only separates head and tail but also \textbf{re-defines} the label space for the convenience of classifiers. So some of the head samples may be tail classes in new label space due to the $\mathcal{L}_{bal}$ constraint.}
	\label{fig:insight}
\end{minipage}
\hfill
\end{figure*}

We can speculate from Fig.~\ref{fig:error-confusion} that if the head and tail classes are ideally separated, the tail-to-head errors and the complexity of imbalanced learning can be significantly reduced.
Thus, we conduct a simple probing to analyze the effect of head-tail separation in Fig.~\ref{fig:separate-class} (detailed in supplementary materials: Supp.~Sec.~1). We separate the test samples into two groups (head: classes with $>$50 samples; tail: the rest classes) before individually classifying the groups. Fig.~\ref{fig:separate-class} shows that accurate head-tail separation can help long-tail learning.
In light of this, we incorporate a \textbf{D}ynamic \textbf{L}abel \textbf{S}pace \textbf{A}djustment (\textbf{DLSA}) method as shown in Fig.~\ref{fig:insight}.
We propose to first confront the \textit{bottleneck} of head-to-tail bias and deal with long-tailed recognition in a \textbf{two-stage} fashion: first adjust the label space and separate the head and tail samples, and then apply classification for the divided groups respectively.
In virtue of the \textit{inductive bias} of deep learning feature models (\eg, pretrained backbone), we could define the new label space as the initial clusters of features and approach \textit{head-tail separation}. 
Moreover, we hope the clusters in the new label space are balanced and contain only a few classes.
We formulate these assumptions on the latent space as: 
1) \textbf{Head-Tail Separability}: the head and tail are separated during the adjustment;
2) \textbf{Cluster Balancedness}: the clusters have balanced sizes;
3) \textbf{Purity}: samples in each cluster are pure, \ie, belong to as few classes as possible.

Specifically, we propose a \textit{plug-and-play} module which can be embedded in most two-stage learning methods~\cite{decouple}. 
We use Gaussian mixture to model the pretrained features and produce clusters. 
Then the samples are divided into two groups and classified independently, where head and tail classes are desired to be separated.
In practice, multiple modules can be linked to a cascade model to progressively separate head and tail classes and reduce the imbalance bias.
Experiments show that the dataset partition and label re-assignment can effectively alleviate label bias.
On top of that, we find backbone models with various inductive biases have different head-tail class separability. 
With DLSA, we can qualitatively evaluate the feature learning of the backbones from the perspective of head-to-tail bias, which can facilitate long-tailed recognition in practice.

Our main contributions are:
1) Proposing a dynamic label space adjustment paradigm for long-tail learning.
2) Our method is plug-and-play and can boost the performances of the state-of-the-arts on widely-adopted benchmarks.
3) Our method can also act as an evaluation of feature model selection to guide long-tailed feature learning and recognition.

\section{Related Work}
\label{sec:related}

\subsection{Long-Tailed Recognition}

Real-world data are long-tailed distributed, which skews machine learning models. 
Numerous tasks face the challenge of long-tailed data, including object detection~\cite{gupta2019lvis}, attribute recognition~\cite{li2020symmetry} and action understanding~\cite{chao2015hico,li2019transferable}.
There is increasing literature trying to alleviate the bias brought by imbalanced data.

Some small adjustments to the model components can alleviate imbalanced learning.
The most intuitive approach is re-balance the \textbf{data distribution}, either by over-sampling the minority~\cite{buda2018systematic,byrd2019effect}, or under-sampling of majority~\cite{japkowicz2002class,he2009learning,buda2018systematic}. 
Data \textbf{augmentation} and \textbf{generation}~\cite{Remix,MetaSAug,chu2020feature} also flattens the long-tailed distribution and helps tail learning.
Re-balancing by \textbf{loss function} adjust the significance of samples on each class~\cite{huang2016learning,huang2019deep,wang2017learning,LDAM,BALMS,samuel2021distributional,sinha2020class}.
Instead of weighting the losses, some methods balance the \textbf{output logits}~\cite{menon2020long,kim2020adjusting,zhong2021improving} after training.

There are also methods to modify the whole training process.
One of the paths is \textbf{knowledge transfer} from head classes to tail classes~\cite{openlongtail,zhang2021balanced,decouple,zhou2020bbn,jamal2020rethinking,kim2020m2m}.
Liu~\etal~\cite{openlongtail} enhance the feature of minority classes with a memory module.
Zhang~\etal~\cite{zhang2021balanced} distill the tail-centric teacher models into the general student network to facilitate tail learning while keeping the head performance.
Some works involve specific balanced \textbf{training strategy} for imbalanced learning.
Kang~\etal~\cite{decouple} exploit the two-stage approach by first training a feature extractor and then re-train a balanced classifier.
Zhou~\etal~\cite{zhou2020bbn} combine uniform sampling and reversed sampling in a curriculum learning fashion.
Jamal~\etal~\cite{jamal2020rethinking} estimate class weights with meta-learning to modulate classification loss.

Recently, self-supervised learning is applied in long-tailed recognition for \textbf{feature space} re-balancing, among which contrastive learning is the most trendy technique~\cite{info-nce,simclr,moco}.
Yang and Xu~\cite{yang2020rethinking} first use self-supervised pretraining to overcome long-tailed label bias.
Kang~\etal~\cite{kang2020exploring} systematically compare contrastive learning with traditional supervised learning and find that the contrastive model learns a more balanced feature space which alleviates the bias brought by data imbalance.
Here, our DLSA provides another strong proof for this conclusion (Sec.~\ref{sec:res}).
Wang~\etal~\cite{wang2021contrastive} combine contrastive feature learning and cross-entropy-based classifier training with a curriculum learning strategy.
Jiang~\etal~\cite{jiang2021self} exploit the contrast of the network and its pruned competitor to overcome the forgetting of the minority class.
Cui~\etal~\cite{paco} propose parametric contrastive learning to rebalance the sampling process in contrastive learning.

However, very few works deal with long-tail bias from the perspective of balancing \textbf{label space}.
Samuel~\etal~\cite{samuel2021generalized-attribute} incorporate semantic class descriptors and enhance minority learning with the familiarity effect of the descriptors. 
Wu~\etal~\cite{wu2020solving-taxonomy} reorganize the label space into tree hierarchy with semantics.
In this paper, we propose DLSA to filter the balance subset from imbalanced data.

\begin{figure*}[!t]
	\begin{center}
		\includegraphics[width=0.97\textwidth]{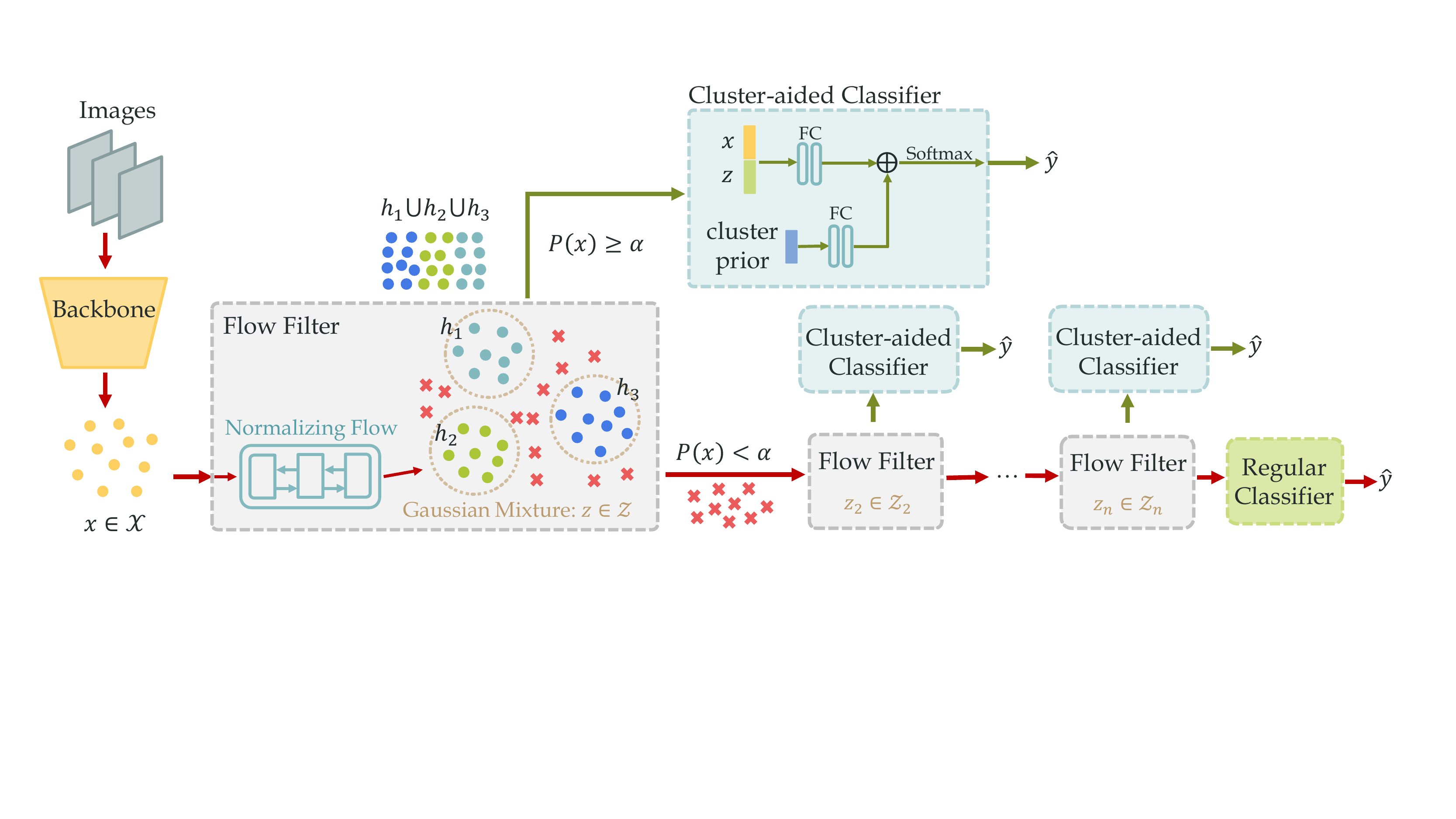}
	\end{center}
	\caption{Method overview. 
	The features $x$ are extracted by pretrained backbones and sent to \textbf{Gaussian Mixture Flow Filter}. The Flow Filter maps the samples to a Gaussian mixture latent space $\mathcal{Z}$ with three constraints: \textbf{cluster balancedness}, \textbf{purity}, and \textbf{head-tail separability} to balance the label space and separate head and tail classes. Then we re-group the data according to the likelihood of each sample. The samples that are more conforming to the Gaussian mixture (dots) are classified by \textbf{Cluster-aided Classifier} with cluster prior information. The samples with low likelihoods (red crosses) are \textit{progressively} solved by Flow Filters or classified by a regular classifier at the last layer. 
	}
	\label{fig:method}
\end{figure*}

\subsection{Normalizing Flows}
\label{sec:rel:nf}
Normalizing flows~\cite{normflow,IAF,MAF,RealNVP} are a family of invertible networks and are widely adopted in density estimation and generative models.
Specifically, a normalizing flow $g$ is a cascaded transforms data $x\in \mathcal{X}$ to representation $u$ in latent space $\mathcal{U}$: $u = g^{-1}(x)$. By assuming a tractable latent distribution, we can recover the data distribution with the \textit{change of variable formula}:
\begin{equation}
    P(x)=P(u)\cdot\left|\mathbf{J}\left(g^{-1}(x)\right)\right|,
\label{eq:chagevar}
\end{equation}
where $\mathbf{J}(\cdot)$ is the Jacobian determinant of a transformation and is a crucial concern in the design of normalizing flows due to its high complexity.
Rezende and Mohamed~\cite{normflow} propose some cascaded normalizing flows and their basic blocks have linear-time Jacobian computation.
Real NVP~\cite{RealNVP} updates only part of the input vector in the normalizing flow block with a simple bijection, which is more capable of modeling high-dimension data distribution. 
IAF~\cite{IAF} and MAF~\cite{MAF} are more general than Real NVP, incorporating autoregressive transformations.
In practical applications of density estimation, the latent distribution $\mathcal{U}$ is usually unit Gaussian for simplicity. Some works~\cite{FlowGMM} exploit Gaussian mixture to explicitly capture the cluster structure of data. In this work, we also use a unit Gaussian mixture to simultaneously transform data and clusters.

\section{Approach} 

\subsection{Preliminaries}

Long-tailed recognition is to learn a model that predict labels $y=1,2,\cdots,\mathcal{C}$ of the data sample $x\in\mathcal{X}$, while the training sample numbers of each class $n_i = \left| \left\{x|y=i,~x\in X_{train}\right\} \right|$ is highly imbalanced. The imbalanceness of data are measured by imbalance factor $\beta=\max(n_i)/\min(n_i)$ and the $\beta$ of typical long-tailed datasets ranges from 10 to 1000.

Since the \textbf{head-tail separation} is a well-known key for long-tail learning, our idea is to separate head and tail samples recursively and classify in a divide-and-conquer manner. So the longtailedness is reduced in each partition of data and, in this sense, we construct \textit{balance from imbalance}.
We first divide head and tail classes, by re-defining a more balanced label space $h=1,2,\cdots,K$.
As discussed, if we can ideally map the data and labels to a balanced space and separate head and tail classes, it will be easier to learn classifiers.
However, it is impractical to achieve an \textit{absolutely} balanced space. So we propose to \textit{progressively} map and cluster samples, and filter out the samples that are well subject to the balanced distributions. During the clustering, we force the model to learn head-tail class separability.
In this paradigm, the complexity of long-tail learning is partially \textit{transferred} to the label space learning and latent space mapping process, while these can be more easily settled with the help of the inductive biases of pretrained features and the transformation ability of normalizing flows~\cite{normflow}.

\noindent{\bf Backbone Models.} Different inductive biases exist in the various deep learning models and training schemes.
Cross-entropy-based supervised learning imposes class separability on deep feature learning.
As previously studied~\cite{decouple}, for deep supervised learning, the data imbalance is mainly detrimental to the classifiers while the feature learning suffers less. Self-supervised contrastive learning methods~\cite{info-nce,simclr} introduce instance-level transformation invariance to deep models, leading to a more balanced feature since there are no labels involved. Further supervised contrastive learning~\cite{yang2020rethinking,kang2020exploring} combines the two aspects and incorporates both instance-level multi-view invariance and class-wise contrast. We compare the main-stream pretrained models with different inductive biases in Sec.~\ref{sec:res} and supervised contrastive learning shows its superiority over the rest.

\noindent{\bf Normalizing Flow.} As introduced in Sec.~\ref{sec:rel:nf}, normalizing flows are invertible transformations and are suitable for density estimation and distribution mapping. We utilize normalizing flows to simultaneously map the data $x$ to representations $z$ in a more balanced latent space and estimate the new class label $h$. The mathematical beauty of normalizing flow enables the maximum likelihood estimation and the learning of data mapping.
We also design multiple constraints for the normalizing flows to re-balance the latent space.

\subsection{Overview}
Fig.~\ref{fig:method} depicts the pipeline of our method. We apply our DLSA in feature space and the initial features are extracted by pretrained self-supervised models.
There are mainly two types of modules arranged in a cascaded manner. \textbf{Gaussian Mixture Flow Filter} (Sec.~\ref{sec:GMFF}) clusters the input data in latent space and assigns new cluster labels.
Then, the samples are divided into two groups. The well-clustered samples are filtered out to mitigate the head-to-tail bias and learning complexity, and are sent to a dedicated \textbf{Cluster-aided Classifier} (Sec.~\ref{sec:CAC}) which exploits the cluster information in classification. 
The rest outliers are either forwarded to the next normalizing flow or regularly classified.

Specifically, the input features are transformed by normalizing flows to a Gaussian mixture distribution to obtain cluster labels. For better separability in the long-tailed scenario, besides the maximum likelihood objective of Gaussian mixture, the flow model is constrained by three additional \textbf{objectives}:
1) Head-tail Separability: head and tail classes should be separated;
2) Cluster Balancedness: the sizes of latent clusters should be as balanced as possible;
3) Purity: the samples in each cluster should belong to as few classes as possible.

\subsection{Gaussian Mixture Flow Filter} 
\label{sec:GMFF}

Gaussian Mixture Flow Filter modules transform data to a desired latent distribution, perform clustering and filter the samples. 
Without loss of generality, we take the first Flow Filter (magnified in Fig.~\ref{fig:method}) as an example.

The filter module incorporates a normalizing flow model $g^{-1}_\theta: \mathcal{X}\rightarrow \mathcal{Z}$ with trainable parameter $\theta$, mapping data samples $x\in\mathcal{X}$ to latent distribution $z\in\mathcal{Z}$. To cluster the input samples into $K$ clusters, we assume the latent distribution $P(z)$ is a Gaussian mixture and each component corresponds to a cluster:

\begin{equation}
    P(z) = \sum_{k=1}^K P(h=k)P(z|h=k),
\end{equation}
where $h$ is the random variable of cluster index. $P(z|h=k) = \mathcal{N}(z|\mu_k, I)$ is a Gaussian probability density with mean vectors $\mu_k$ and identity covariance matrix. The $\mu_k$ are randomly sampled from a normal distribution and fixed during training~\cite{FlowGMM}. 
A proper assumption on \textit{prior probability} $P(h)$ is required for a tractable optimization methods of NF with GMM.
Since $P(h)$ in GMM determines the ``size'' of a Gaussian component, with the \textbf{balancedness} assumptions, the prior probability $P(h)$ can be set to uniform distribution, thus:
\begin{equation}
    P(z) = \frac1K \sum_{k=1}^K P(z|h=k),
\end{equation}
and the prediction of sample $x$ is given by Bayes' theorem:
\begin{equation}
    P(k|x) = \frac{P(z|h=k)}{\sum_{k'=1}^K P(z|h=k')}, \quad
    \hat{h} = \argmax{k} P(k|x).
\end{equation}

We train the normalizing flow model with end-to-end gradient descent optimization with the following objectives.

\noindent{\bf Maximum Likelihood Loss and Head-tail Separability.}
We propose a weighted maximum likelihood loss to simultaneously train the cluster model and learn head-tail class separation.
With the elegant invertible property of normalizing flow and change of variable formula Eq.~\ref{eq:chagevar}, we can painlessly obtain the Gaussian mixture likelihood of data sample $x$:
\begin{equation}
    L(\theta;x) = \sum_{k=1}^K \frac1K \mathcal{N}(g^{-1}_{\theta}(x)|\mu_k, I) \left| \mathbf{J}\left( g^{-1}_{\theta}(x) \right) \right|.
\end{equation}
The maximum likelihood estimation (MLE) of parameter $\theta$ will produce a clustering model. 
To enhance the separability of head and tail classes of the Flow Filter, we incorporate \textit{sample weighting} and impose higher weights on tail samples than head samples. In DLSA, samples in class $i$ are weighted by:
\begin{equation}
    \omega(i)=\frac{n_i^{-q}}{\sum_j n_j^{-q}},
\end{equation}
where $n_i$ is the training sample size of class $i$. $q$ is a positive number usually ranging from 1 to 2, hence the tail samples receive larger weights. 
The weighted $L(\theta;x)$ will impose higher likelihoods on tail samples than head samples, therefore tail classes are more likely to be filtered out by Flow Filter and separated with head classes. 
The weighted maximum likelihood loss is constructed with the negative log-likelihood of a data batch $\mathcal{B}$:
\begin{equation}
    \mathcal{L}_{\mathit{MLE}}(\mathcal{B}) = -\sum_{x\in \mathcal{B}}\omega(y)\log L(\theta; x).
\end{equation}

\begin{figure*}[!t]
\begin{minipage}{0.45\textwidth}
	\centering
	\includegraphics[width=0.95\textwidth]{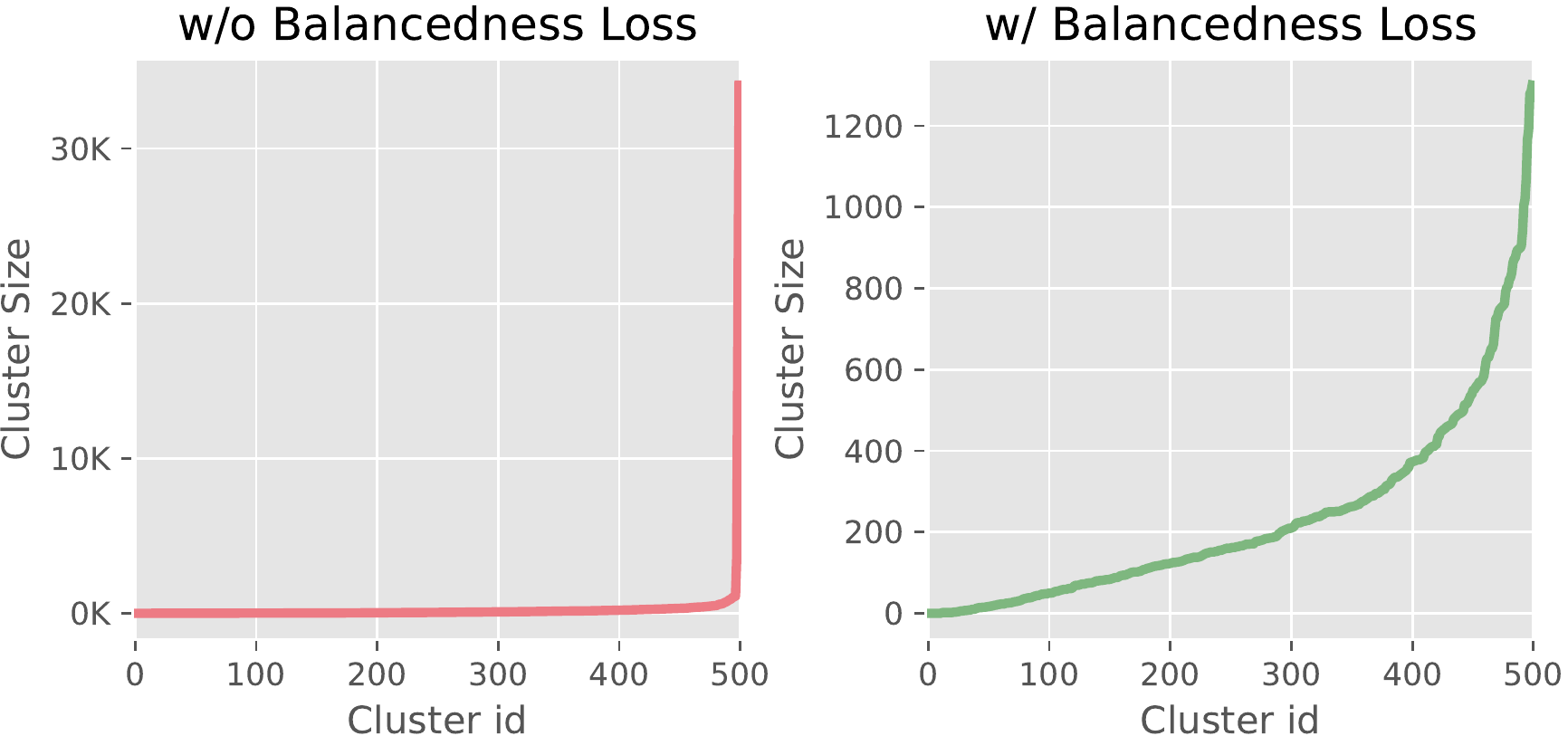}
	\caption{Cluster sizes of DLSA w/ or w/o $\mathcal{L}_{bal}$ on ImageNet-LT~\cite{decouple}.}
	\label{fig:balance-cmp}
\end{minipage}
\hfill
\begin{minipage}{0.48\textwidth}
	\centering
	\includegraphics[width=0.95\textwidth]{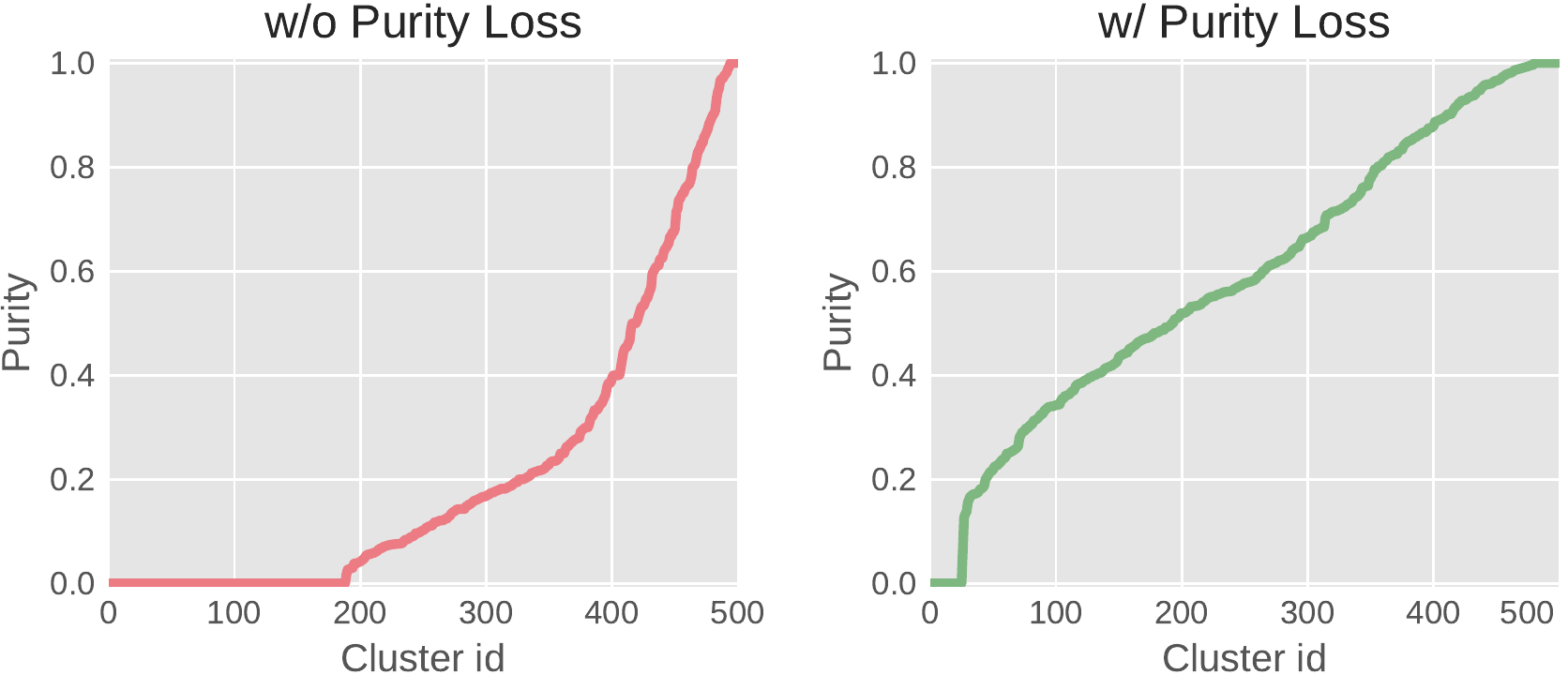}
	\caption{Cluster purity of DLSA w/ or w/o $\mathcal{L}_{pure}$ on ImageNet-LT~\cite{decouple}.}
	\label{fig:purity-cmp}
\end{minipage}

\end{figure*}

Training the flow model with a single MLE constraint may result in trivial solutions or a latent distribution that is not suitable for long-tailed recognition, \eg, $g^{-1}_{\theta}$ maps all samples to one cluster (Fig.~\ref{fig:balance-cmp} and \ref{fig:purity-cmp}). To avoid \textit{model collapse} and \textit{degeneration}, we introduce two more objectives for the flow filter.

\noindent{\bf Cluster Balancedness Loss.}
Though the prior distribution $P(h)$ has been set to uniform, the posterior after observing data samples $X$:
\begin{equation}
    P(h=k|X)  = \sum_{x\in X}P(x|X)P(h=k|x)
                   = \sum_{x\in X}\frac1{|X|}P(h=k|x),
\end{equation}
can still be imbalanced; if so, the actual cluster sizes would be biased or even severely long-tailed distributed. To promote cluster balance, we introduce
\begin{equation}
\mathcal{L}_{bal}(X) = \mathbb{E}_h\left[ \log P(h|X) \right],
\end{equation}
a cluster balancedness loss which is the negative entropy of the posterior. It is also equivalent to the KL divergence between $P(k|X)$ and a discrete uniform distribution, whose simple proof is shown in Supp.~Sec.~2.1. So minimizing the loss will force the cluster sizes to be more even. As shown in Fig.~\ref{fig:balance-cmp}, balancedness loss can significantly reduce the unbalancedness of the cluster sizes.

In practice, the optimization of $\mathcal{L}_{bal}$ can be unstable, especially with mini-batch based optimizer~\cite{kiefer1952stochastic}. For each mini-batch $\mathcal{B}_t$ at $t^{th}$ iteration, $\hat{p}=P(h=k|\mathcal{B}_t)$ is an unbiased estimator of the posterior $P(h=k|X)$. But small batch size can lead to large variance of $\hat{p}$. Therefore, we propose to exploit the \textit{history information} and incorporate a momentum-based estimator: $\tilde{p}_t = \eta\tilde{p}_{t-1}+(1-\eta)P(h=k|\mathcal{B}_t)$,
where $\eta$ is decay factor. $\tilde{p}_{t-1}$ is the estimation of last batch and we set $\tilde{p}_{0}$ to $0$. Thus, the bias-corrected estimator $\frac{\tilde{p}_t}{1-\eta^t}$ is an unbiased and more efficient estimator than $\hat{p}$, and is more stable during mini-batch training. The proof of its unbiasedness and efficiency is shown in Supp.~Sec.~2.2.

\noindent{\bf Purity Loss.}
We define a scalar ``purity'' $\mathit{Purity}(k)$ as the sample proportion of the largest class in the cluster $k$.
To enhance the class purity in each clusters, we repeatedly randomly sample a pair of samples $x_i$, $x_j$ from two different classes $i$, $j$ and suppressing their similarity in cluster predictions:
\begin{equation}
\mathcal{L}_{pure}(x_i, x_j) = \sum_{k=1}^K P(h=k|x_i)\log P(h=k|x_j).
\end{equation}
By minimizing the purity loss, we increase the cross-entropy between $P(h|x_i)$ and $P(h|x_j)$, thus pushing the two samples farther. Fig.~\ref{fig:purity-cmp} indicates purity loss effectively increases the purity of the clusters.
Different sampling strategies can be applied to the purity loss. Our experiments show that the best practice is sampling with equal probability for each class.

Overall, the total loss of Flow Filter module is:
\begin{equation}
\mathcal{L}_{total} = \mathcal{L}_{\mathit{MLE}} + \lambda_{bal}\mathcal{L}_{bal} + 
                        \lambda_{pure}\mathcal{L}_{pure}.
\end{equation}
Among the constraints, $\mathcal{L}_{MLE}$ is for learning the data distribution and its sample-wise weights make the model lean to head samples, enabling head-tail separation with sample likelihood $P(x)$. 
$\mathcal{L}_{pure}$ ensures that each cluster in GMM belongs to only one class, so we can use cluster information to enhance classification. 
$\mathcal{L}_{bal}$ reduces the longtailedness in clusters and prevents the model from trivial solutions, \eg, most samples go to one cluster.
$\mathcal{L}_{pure}$ and $\mathcal{L}_{bal}$ ensure the training stability and enhance the separatability learned by $\mathcal{L}_{MLE}$.

\subsubsection{Filtering}

The Flow Filter implements the divide step in the DLSA pipeline.
After training, some samples do not belong to any cluster and become outliers of the Gaussian mixture. We use a likelihood threshold $\alpha$ to separate the well-clustered samples and outliers. As shown in Fig.~\ref{fig:method}, the samples with higher confidence ($P(x)\leq \alpha$, green and blue) are closer to the cluster centers and the rest ($P(x)> \alpha$, red) are outliers of any clusters so they are sent to next adjustment module to further decrease the training difficulty. Accordingly, the samples are divided into two groups to separate head and tail classes.

In practice, we set the threshold $\alpha$ to a quantile of $P(x)$ on training data, so that a certain proportion of data will be filtered out.

\subsection{Cluster-Aided Classifier}
\label{sec:CAC}

The high-confidence samples filtered out by the Flow Filter are sent to a Cluster-aided Classifier. Under ideal conditions, each cluster $h$ in the Flow Filter contains only one class $y$, so the prediction can be directly obtained by a simple label-class mapping.
However, the learned Gaussian mixture is rather noisy and besides a majority class $y$ in one cluster, there are some samples from other classes. To compensate for these noise samples, we introduce a softer label-class mapping method. As demonstrated in Fig.~\ref{fig:method}, for each sample $x_i$, we first compute cluster prior $P(y|h=\hat{k}_i)$, which is the class frequency of training samples belonging to the sample's predicted cluster $\hat{k}_i$. The cluster prior vector and the concatenation of feature $x_i$ and latent representation $z_i$ are forwarded to two independent fully-connected layers (FC). Their outputs are added and we take its Softmax as output probability: $P(y|x,z) = \mathit{Softmax}\left(FC[x,z]+FC[P(y|h=\hat{k}_i)]\right)$.

\section{Experiment}

\subsection{Datasets}
We evaluate our approach on three main-stream benchmarks: ImageNet-LT~\cite{openlongtail}, Places-LT~\cite{openlongtail}, and iNaturalist18~\cite{van2018inaturalist}. 
ImageNet-LT~\cite{openlongtail} and Places-LT~\cite{openlongtail} are long-tailed subsets of ImageNet~\cite{imagenet} and Places-365~\cite{zhou2017places}, with 1,000 and 365 categories and about 186 K and 106 K images respectively. Both datasets have Pareto distributed train sets and \textit{balanced test sets} and the imbalanced factor $\beta$ is 256 and 996.
iNaturalist18~\cite{van2018inaturalist} is a fine-grained image classification dataset which is naturally highly imbalanced ($\beta=500$), with 8,142 categories and over 437 K images.
Following \cite{openlongtail}, we report overall accuracy on all datasets, and Many-shot (classes with over 100 images), Medium-shot (classes with 20-100 images), Few-shot (classes with fewer than 20 images) accuracy. 
We also use Matthews correlation coefficient (MCC)~\cite{matthews1975comparison-mcc} and normalized mutual information (NMI)~\cite{danon2005comparing-nmi} to measure the performance in long-tailed scenario.

\subsection{Baselines}

We embed and evaluate our models in multiple two-stage~\cite{decouple} long-tail learning methods, with different trending feature learning and classifier learning. The involved methods cover various long-tail learning strategies, including data rebalancing, loss balancing, feature space balancing, and decision boundary adjusting.

\noindent{\bf Feature Learning.} 
(1) \textbf{Supervised Cross-Entropy (CE)}: a baseline backbone trained with vanilla cross-entropy loss. 
(2) \textbf{Class-balance Sampler} \textbf{(CBS)} \cite{CBSampler}: a balanced data sampler where each class has equal sampling probability.
(3) \textbf{PaCo}~\cite{paco}: supervised contrastive learning model based on MoCo~\cite{moco}
It manages to alleviate the sampling bias in contrastive learning. PaCo is currently the state-of-the-art approach on the two mentioned benchmarks. Following the original implementation, it is integrated with Balanced Softmax~\cite{BALMS}. We reproduce the two-stage training version of PaCo and RandAug~\cite{cubuk2020randaugment} is not applied in classifier learning for a fair comparison with other features.

\noindent{\bf Classifier Learning.} 
(1) \textbf{BalSoftmax}~\cite{BALMS}: an adjusted Softmax-cross-entropy loss for imbalanced learning, by adding $\log(n_j)$ to logit of class $j$ during training, where $n_j$ is the training sample size of class $j$.
(2) \textbf{Class-balance Sampler (CBS)}~\cite{CBSampler}: same as CBS in Feature Learning.
(3) \textbf{Cosine classifier (Cosine)}~\cite{cosine1,cosine2}: predicting according to the cosine distance of features $x$ and class embeddings $w_i$. Or namely, normalize the features $x$ and weight vector $w_i$ of each class $i$ of a linear classifier:
$\hat{y}=\text{argmin}_i \{\cos \langle w_i,~x\rangle \} $
(4) \textbf{M2M}~\cite{kim2020m2m} transfers head information to tail by resampling strategies.
(5) \textbf{RIDE}~\cite{wang2020RIDE} trains multiple diversified expert classifier and dynamically select the best expert for each sample. We use RIDE with two experts.

\subsection{Implementation Details}

We use 2-layered MAF~\cite{MAF} as the Flow Filter model. The Gaussian mixture centers are randomly sampled from $\mathcal{N}(0, 0.05^2)$ and the details of variance selection please refer to Supp.~Sec.~3.
The model is trained with a SGD optimizer and the decay rate for balancedness loss momentum $\eta=0.7$.
By default, the weight $q$ in the $\mathcal{L}_{MLE}$ is 2.0, and $30\%$ samples are filtered to Cluster-aided Classifiers in each division step. \textbf{We reproduce the baselines with decoupled strategy}~\cite{decouple}.
The detailed hyper-parameters of each dataset are listed in Supp.~Sec.~5.

\subsection{Results}
\label{sec:res}

\begin{table}[t]
\centering
\small
\caption{\small Results on ImageNet-LT~\cite{openlongtail} with ResNet-50~\cite{resnet} backbone.}
\adjustbox{width=0.75\linewidth}{
\begin{tabular}{l|l|cccc|cc}
	\toprule
	Feature & Classifier & Overall & Many & Medium & Few & MCC & MNI \\
	\midrule
	\multirow{2}{*}{CE}
	        & BalSoftmax~\cite{BALMS}          & 42.8 & 54.1 & 39.4 & 23.2 &  42.7  & 70.1 \\
	        & BalSoftmax~\cite{BALMS} + DLSA   & \textbf{43.9} (+\textbf{1.1}) & \textbf{54.5} & \textbf{41.0} & \textbf{24.0} & \textbf{43.8} & \textbf{70.3}\\
	        
	\midrule
	\multirow{2}{*}{CBS~\cite{CBSampler}}
	        & BalSoftmax~\cite{BALMS}          & 42.2 & \textbf{55.8} & 38.3 & 17.6 & 42.2  & 69.9\\
	        & BalSoftmax~\cite{BALMS} + DLSA   & \textbf{43.1} (+\textbf{0.9}) & 55.3 & \textbf{40.2} & \textbf{18.9} & \textbf{43.1} & \textbf{70.2} \\

	\midrule
	\multirow{10}{*}{PaCo~\cite{paco}}
	        & CBS~\cite{CBSampler}             & 54.4 & 61.7 & 52.0 & 42.5 &  54.4  & 74.5\\
	        & CBS~\cite{CBSampler} + DLSA      & \textbf{55.6} (+\textbf{1.2}) & \textbf{62.9} & \textbf{52.7} & \textbf{45.1} & \textbf{55.6} & \textbf{74.9} \\
	\cmidrule{2-8}
	        & BalSoftmax~\cite{BALMS}         & 54.9 & 67.0 & 50.1 & 38.0 &  54.9  & 74.9\\
	        & BalSoftmax~\cite{BALMS}+ DLSA   & \textbf{56.3} (+\textbf{1.4}) & \textbf{67.2} & \textbf{52.1} & \textbf{40.2} & \textbf{56.1} & \textbf{75.4}\\
	\cmidrule{2-8}
	        & Cosine~\cite{cosine1,cosine2}          & 55.7  & \textbf{64.9}  & 53.0  & 39.5 &  55.7  & 75.2\\
	        & Cosine~\cite{cosine1,cosine2} + DLSA   & \textbf{56.9} (+\textbf{1.2}) &  64.6 & \textbf{54.9} & \textbf{41.8} & \textbf{56.8}  & \textbf{75.7} \\
	\cmidrule{2-8}
	        & M2M~\cite{kim2020m2m}          & 55.8 & 67.3 & 52.1 & 36.5 & 55.8 & 75.3\\
	        & M2M~\cite{kim2020m2m} + DLSA   & \textbf{56.7} (+\textbf{0.9}) & \textbf{68.0} & \textbf{52.8} & \textbf{38.2} & \textbf{56.6} & \textbf{75.7}\\
	\cmidrule{2-8}
	        & RIDE~\cite{wang2020RIDE}          & 56.5 & 67.3 & 53.3 & 37.3 & 56.5 & 75.6\\
	        & RIDE~\cite{wang2020RIDE} + DLSA   & \textbf{57.5} (+\textbf{1.0}) & \textbf{67.8} & \textbf{54.5} & \textbf{38.8} & \textbf{57.5} & \textbf{75.9}\\
	        
	\bottomrule
\end{tabular}}
\label{tab:res-imgnet}
\end{table}

We evaluate the methods on some baselines in Tab.~\ref{tab:res-imgnet}, \ref{tab:res-places} and \ref{tab:res-inat}. 
Supp.~Sec.~6 shows the detailed results on iNaturalist18.
Our method promotes the accuracy of all the baselines. On state-of-art method PaCo~\cite{paco}+BalSoftmax~\cite{BALMS}, we achieve accuracy gain of 1.4\%, 1.2\%, 1.0\% on the datasets respectively. Especially, on few-shot classes, the method brings over 1.5\% accuracy bonus.
Though iNaturalist18 is a large-scale, realistic, and severely imbalanced dataset, DLSA still brings 1\% improvement with PaCo representations.
DLSA also brings about 1\% and 0.5\% gain on MCC and NMI score.

\noindent{\bf Comparison of classifiers.}
In Tab.~\ref{tab:res-imgnet} and \ref{tab:res-places}, we compare different classifier learning models on same pretrained features (last 10 rows of Tab.~\ref{tab:res-imgnet} and rows beginning with ``CE+'' in Tab.~\ref{tab:res-places}). On each dataset, all the classifier methods use a same trained Flow Filter model since we incorporate two-stage training in DLSA and the learning of Flow Filters only depends on the feature models. DLSA brings comparable improvement on different classifiers (1.2\%, 1.4\%, 1.2\%, 0.9\%, 1.0\% on ImageNet-LT; 1.2\%, 1.2\%, 1.0\% on Places-LT). The performance improvement of our model mainly depends on the initial structure of the features.

\begin{wrapfigure}{r}{0.35\textwidth}
\includegraphics[width=0.34\textwidth]{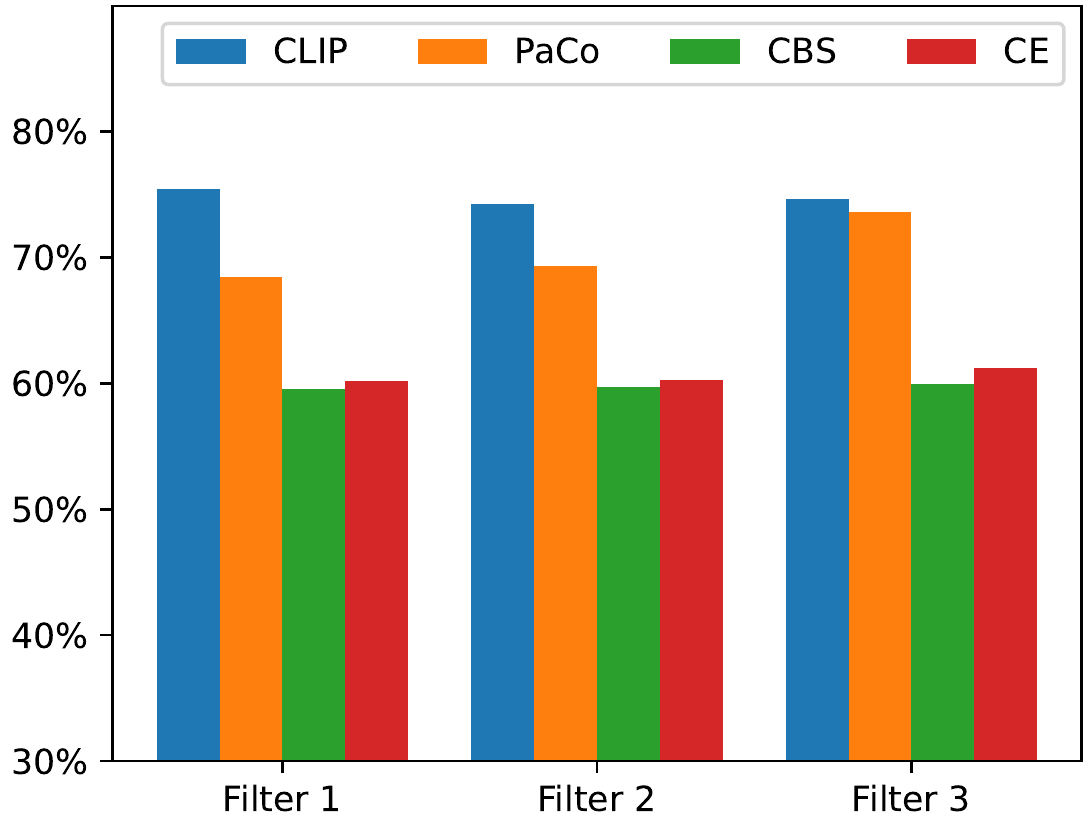}
\caption{\small The accuracy of head-tail separation of DLSA on different feature models. The threshold of head/tail is 50.}
\label{fig:backbone-sep}
\end{wrapfigure}
\subsubsection{Feature model comparison and evaluation.}
With DLSA, models with different feature models have comparable overall accuracy improvement (1.1\%, 0.9\%, 1.4\% with BalSoftmax classifier on ImageNet-LT), but the improvement on few-shot classes differs (0.8\%, 1.3\%, 2.2\% with BalSoftmax classifier). This is due to the different \textit{head-tail separability} of pretrained features with different inductive biases. 
Thus, we compare the head-tail separation accuracy of DLSA on the three feature models.
We analyze the samples filtered out by Flow Filters and compute the accuracy of these samples being tail classes. As discussed in Sec.~\ref{sec:intro}, an ideal separation with 100\% accuracy will significantly boost the long-tailed recognition. On PaCo+BalSoftmax, 100\% accurate head-tail separation brings $\sim$10\% improvement.
Fig.~\ref{fig:backbone-sep} shows the accuracy of head-tail classes separation of models with different features. At all three Flow Filter layers, features of PaCo show superior separability than feature models pretrained with CE or CBS, thus PaCo has more potential of enhancing tail recognition. We also evaluate the DLSA on CLIP~\cite{CLIP} features of ImageNet-LT. It achieves \textbf{65.2}\% overall accuracy and over \textbf{75}\% separation accuracy, surpassing all baselines, showing the \textbf{potential} of DLSA in the latest feature spaces.
These also strongly support the conclusion of \cite{kang2020exploring}, that contrastive learning generates a more balanced feature space and enhances generalization capability.
The positive correlation between head-tail separability and performance gain indicates that our proposed DLSA is a practical tool for feature analysis and evaluation with long-tailed data.

\begin{table}[t]
\centering
\caption{\small 
    Results on Places-LT~\cite{openlongtail} with ImageNet~\cite{imagenet}-pretrained ResNet-152~\cite{resnet}.
}

\adjustbox{width=0.65\linewidth}{
\begin{tabular}{l|cccc|cc}
	\toprule
	Feature & Overall & Many & Medium & Few & MCC & NMI \\
	\midrule
	Joint (baseline)~\cite{decouple}  & 30.2 & 45.7 & 27.3 &  8.2 & - & - \\
	cRT~\cite{decouple}               & 36.7 & 42.0 & 37.6 & 24.9 & - & - \\
    $\tau$-norm~\cite{decouple}       & 37.9 & 37.8 & 40.7 & 31.8 & - & - \\
    LWS~\cite{decouple}               & 37.6 & 40.6 & 39.1 & 28.6 & - & - \\
    MetaDA~\cite{MetaDA}           & 37.1 &  -   &  -   &  -   & - & - \\
    LFME~\cite{LFME}               & 36.2 & 39.3 & 39.6 & 24.2 & - & - \\
	FeatAug~\cite{chu2020feature}  & 36.4 & 42.8 & 37.5 & 22.7 & - & - \\
	BALMS~\cite{BALMS}             & 38.7 & 41.2 & 39.8 & 31.6 & - & - \\
	RSG~\cite{wang2021rsg}         & 39.3 & 41.9 & 41.4 & 32.0 & - & - \\
	DisAlign~\cite{DisAlign}       & 39.3 & 40.4 & 42.4 & 30.1 & - & - \\
	\midrule
	CE+BalSoftmax~\cite{BALMS}        & 37.8 & 40.0 & \textbf{40.2} & 28.5 & 37.7 & 58.1 \\
	CE+BalSoftmax~\cite{BALMS}+ DLSA  & \textbf{39.0} (+\textbf{1.2}) & \textbf{42.0} & 39.8 & \textbf{31.3} & \textbf{38.8} & \textbf{58.6} \\
	\midrule
	CE+Cosine~\cite{cosine1,cosine2}        &  37.0 & 39.2 & 37.7 & 31.4 & 36.8 & 57.8 \\
	CE+Cosine~\cite{cosine1,cosine2}+ DLSA  &  \textbf{38.2} (+\textbf{1.2}) & \textbf{39.7} & \textbf{39.2} & \textbf{33.3} & \textbf{38.1} & \textbf{58.3} \\
	\midrule
	CE+RIDE~\cite{wang2020RIDE}        & 41.2 & 44.4 & \textbf{43.2} & 31.1 & 41.1 & 59.7 \\
	CE+RIDE~\cite{wang2020RIDE}+ DLSA  & \textbf{42.2} (+\textbf{1.0}) & \textbf{45.8} & 43.0 & \textbf{33.7} & \textbf{42.0} & \textbf{60.2} \\
	\midrule
	PaCo+BalSoftmax~\cite{paco}               & 40.9 & \textbf{44.5} & 42.7 & 30.4 & 40.8 & 59.6 \\
	PaCo+BalSoftmax~\cite{paco}+ DLSA         & \textbf{42.1} (+\textbf{1.2}) & 44.4 & \textbf{44.6} & \textbf{32.3} & \textbf{42.0} & \textbf{59.7} \\
	\bottomrule
\end{tabular}}
\label{tab:res-places}
\end{table}

\begin{figure*}[t]
\centering
\includegraphics[width=0.95\textwidth]{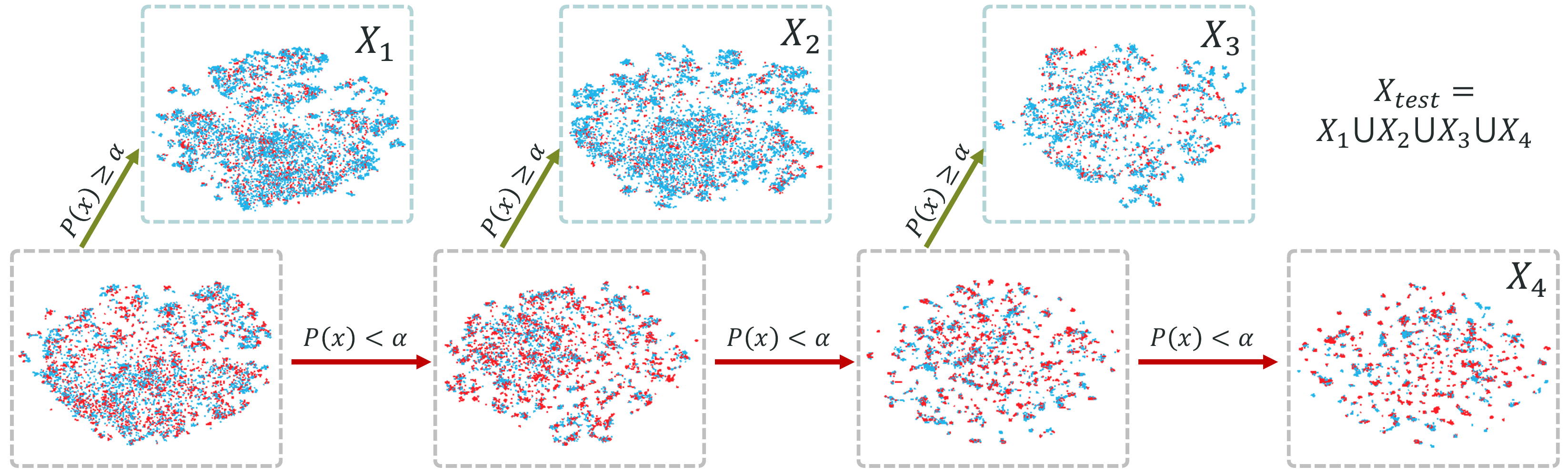}
\caption{Sample separation process of PaCo~\cite{paco}+BalSoftmax~\cite{BALMS} on ImageNet-LT~\cite{openlongtail} test set. The \textbf{\textcolor{red}{red}} and \textbf{\textcolor{blue}{blue}} points are \textbf{\textcolor{red}{head}} and  \textbf{\textcolor{blue}{tail}} samples respectively. The grey boxes are Flow Filters and blue boxes are Cluster-aided Classifiers.}
\label{fig:visualize}
\end{figure*}

\subsection{Visualization}
We visualize our cascade division process in Fig.~\ref{fig:visualize} via t-SNE~\cite{tsne}. We take the PaCo+BalSoftmax as the example.
Our method can effectively divide the samples of head and tail classes gradually. The samples filtered out to Cluster-aided Classifiers (in the blue rectangles) are mainly tail samples, while the samples passed to the next Flow Filters (in the grey rectangles) are mixed.

\begin{table*}[t]

\begin{minipage}{0.45\textwidth}
    \centering
    \caption{\small Results on iNaturalist-18~\cite{van2018inaturalist} with ResNet-50~\cite{resnet}.} 
    \adjustbox{width=0.6\linewidth}{
    \begin{tabular}{l|c}
    	\toprule
    	Feature & Overall \\
    	\midrule
        LWS~\cite{decouple}            & 69.5 \\
        MetaDA~\cite{MetaDA}           & 67.6 \\
        Deep-RTC~\cite{wu2020solving-taxonomy} & 64.0 \\
        Remix~\cite{Remix}             & 70.5 \\
    	FeatAug~\cite{chu2020feature}  & 65.9 \\
    	smDRAGON~\cite{DRAGON}         & 69.1 \\
    	KCL~\cite{kang2020exploring}   & 68.6 \\
    	MiSLAS~\cite{MiSLAS}           & 71.6 \\
    	DisAlign~\cite{DisAlign}       & 70.6 \\
    	Hybrid-PSC~\cite{wang2021contrastive} & 70.4 \\
    	FSR~\cite{FSR}                 & 65.5 \\
    	DRO-LT~\cite{DRO-LT}           & 69.7 \\
    	DiVE~\cite{DiVE}               & 71.7 \\
    	\hline
    	PaCo~\cite{paco}               & 71.8 \\
        PaCo~\cite{paco} + Ours        & \textbf{72.8} (+\textbf{1.0}) \\
    	\bottomrule
    \end{tabular}}
    \label{tab:res-inat}
\end{minipage}
\hfill
\begin{minipage}{0.5\textwidth}
    \centering
    \caption{Ablation study on ImageNet-LT~\cite{openlongtail} with PaCo~\cite{paco}+BalSoftmax~\cite{BALMS} model and ResNet-50~\cite{resnet} backbone.}
    \adjustbox{width=0.9\linewidth}{
    \begin{tabular}{l|l|cccc}
    	\toprule
    	Method & Overall & Many & Medium & Few \\
    	\midrule
    	Full model      & \textbf{56.3} & \textbf{67.2} & \textbf{52.1} & \textbf{40.2} \\
    	\midrule
    	w/o $\mathcal{L}_{\mathit{MLE}}$   & 54.7 & 66.1 & 50.8 & 36.3 \\
    	w/o $\mathcal{L}_{bal}$   & 55.2 & 65.8 & 51.2 & 38.9 \\
    	w/o $\mathcal{L}_{pure}$  & 55.4 & 66.1 & 51.5 & 39.2 \\
    	\midrule
    	300 clusters    & 55.5 & 66.1 & 51.3 & 39.3 \\
    	1000 clusters   & 55.1 & 66.5 & 50.3 & 39.6 \\
    	\midrule
    	2 blocks   & 55.4 & 66.3 & 51.2 & 39.3 \\
    	4 blocks   & 55.8 & 66.5 & 51.7 & 39.2 \\
    	\bottomrule
    \end{tabular}}
    \label{tab:abl}
\end{minipage}

\end{table*}

\subsection{Ablation Study}
\label{sec:abl}

We conduct ablation studies and look into the model components on the test set of ImageNet-LT~\cite{openlongtail} with PaCo~\cite{paco}+BalSoftmax~\cite{BALMS} model, from the following aspects. 
The results are reported in Tab.~\ref{tab:abl}.
The extended ablation studies on Places-LT are shown in Supp.~Sec.~7.

\noindent{\bf Objectives.} We evaluate the three objectives $\mathcal{L}_{\mathit{MLE}}$, $\mathcal{L}_{bal}$, $\mathcal{L}_{pure}$ by removing one of them. Removing any loss leads to a significant performance drop, which indicates the necessity of all objectives. Among these losses, w/o $\mathcal{L}_{pure}$ shows the least degradation since the pretrained backbones have moderately learned intra-class similarity and $\mathcal{L}_{pure}$ aim to explicitly enhance the compactness.

\noindent{\bf Cluster number.} Models with less/more clusters perform worse than default 500 clusters. Larger cluster number results in slow training and inference too.

\noindent{\bf Filter block number.} Models with more Flow Filter blocks can separate head and tail classes more finely. But excessive division operations lead to few training samples for the clustering model and classifiers. Due to this trade-off, the default model with 3 blocks outperforms that with 2 or 4 blocks.

\section{Conclusions}

In this paper, we propose to confront the head-to-tail bias by re-balancing the label space and separating head and tail classes. We present a plug-and-play module DLSA, which automatically adjusts the data distribution and constructs new label space to facilitate the recognition.
We embed DLSA in various types of long-tailed recognition state-of-the-arts and boost their performances. 
We observe that DLSA is also capable of evaluating the different feature learning models.
Our future work may extend to combining our paradigm with end-to-end models and more tasks, \eg, object detection and segmentation.

\noindent{\bf Acknowledgement} This work was supported by the National Key R\&D Program of China (No. 2021ZD0110700), Shanghai Municipal Science and Technology Major Project (2021SHZDZX0102), Shanghai Qi Zhi Institute, and SHEITC (2018-RGZN-02046).

\clearpage

\title{Supplementary Material for Constructing Balance from Imbalance for Long-tailed Image Recognition} 

\titlerunning{Supplementary for Constructing Balance from Imbalance}

\authorrunning{Y. Xu et al.}

\author{}
\institute{}

\maketitle

\section{The Settings of Fig. 2 experiments}

The separation model randomly sends samples to two groups with different probabilities: $(p, 1-p)$ for head classes and $(1-p, p)$ for tail classes. So larger $p$ leads to higher accuracy. The classifier is a ResNet-50 trained with Adam optimizer Tail with lr=3e-4 and bz=512.

\section{Proofs in Cluster Balancedness Loss}
\subsection{Equivalence of KL Divergence and Negative Entropy.}
The KL divergence of $P(h|X)$ and discrete uniform distribution $u(h)$ is:

\begin{equation}
\begin{aligned}
& KL\left[P(h|X)\|u(h) \right] \\
=& \sum_{k=1}^{K}P(h=k|X)\log\frac{P(h=k|X)}{u(h=k)} \\
=& \mathbb{E}_h\left[ \log P(h|X) \right] - \sum_{k=1}^{K}P(h=k|X)\log{\frac1K} \\
=& -Ent\left[P(h|X)\right] + \log K.
\end{aligned}
\end{equation}

$Ent\left[P(h|X)\right]$ is the entropy of $P(h|X)$.
Since $\log K$ is a constant, the optimization of KL divergence and negative entropy are equivalent in the \textit{Cluster Balancedness Loss}.

\subsection{Unbiasedness and Efficiency of Momentum Estimator.}

We first convert the recursive formula of momentum estimator to closed-form:
\begin{equation}
\tilde{p}_t = 
\sum_{i=1}^{t}(1-\eta)\eta^{t-i} P(h=k|\mathcal{B}_i).
\end{equation}

(1) Unbiasedness: Since $\hat{p}=P(h=k|\mathcal{B}_t)$ is unbiased $\mathbb{E}\left[\hat{p}\right]=P(h=k|X)$. Therefore we have

\begin{equation}
\begin{aligned}
& \mathbb{E}\left[\frac{\tilde{p}_t}{1-\eta^t}\right] \\
=& \frac{1}{1-\eta^t}\mathbb{E}\left[\sum_{i=1}^{t}(1-\eta)\eta^{t-i} P(h=k|\mathcal{B}_i)\right] \\
=& \frac{1}{1-\eta^t}\sum_{i=1}^{t}(1-\eta)\eta^{t-i}P(h=k|X)  \\
=& P(h=k|X)
\end{aligned}
\end{equation}

is unbiased.

(2) Efficiency:

\begin{equation}
\begin{aligned}
& \mathbb{D}\left[\frac{\tilde{p}_t}{1-\eta^t}\right] \\
=& \frac{(1-\eta)^2}{(1-\eta^t)^2}\mathbb{D}\left[\sum_{i=1}^{t}\eta^{t-i} P(h=k|\mathcal{B}_i)\right] \\
=& \frac{(1-\eta)^2}{(1-\eta^t)^2} \sum_{i=1}^{t}\eta^{2(t-i)} \mathbb{D}\left[\hat{p}\right] \\
=& \frac{(1-\eta)^2}{(1-\eta^t)^2} \cdot \frac{1-\eta^{2t}}{1-\eta^2} \mathbb{D}\left[\hat{p}\right] \\
=& \frac{(1-\eta)/(1+\eta)}{(1-\eta^t)/(1+\eta^t)} \mathbb{D}\left[\hat{p}\right]. \\
\end{aligned}
\end{equation}

Let $\phi(t)=\frac{1-\eta^t}{1+\eta^t}$, which is monotonic increasing when $0<\eta<1$. So $\mathbb{D}\left[\frac{\tilde{p}_t}{1-\eta^t}\right]=\frac{\phi(1)}{\phi(t)}\mathbb{D}\left[\hat{p}\right]\leq \mathbb{D}\left[\hat{p}\right]$ and $\frac{\tilde{p}_t}{1-\eta^t}$ is more efficient than $\hat p$.

\section{Variance of Gaussian Mixture Centers}

Following \cite{FlowGMM}, the components of Gaussian Mixture are $\mathcal{N}(\mu_k,I)$, and each dimension of the center $\mu_k$ is sampled from $\mathcal{N}(0,\sigma^2)$. The $\sigma$ is selected according to the distances between the generated centers. The clusters can overlap if the centers are too close, and samples may be stuck in the low-density area if the centers are far. The mean distance of two centers $\mu_p)$, $\mu_q$ is:

\begin{equation}
\mathbb{E}\left(\|\mu_p-\mu_q\|^2\right) = \sum_{i=1}^D \mathbb{E}\left(\|\mu_{p,i}-\mu_{q,i}\|^2\right) = 2D\sigma^2.
\end{equation}

If we expect the $\mu_p$ distributing around the \textit{three-sigma borders} of $\mu_q$ ($mu_p-\mu_p=3$), then $\sigma=\sqrt{\frac3{2D}}\approx 0.04$ when the feature dimensionality is 1024.
After experiments, we use $\sigma=0.05$ as the best choice.

\section{Confusion Matrices of methods with DLSA}

\begin{figure*}[ht]
\begin{minipage}{0.3\textwidth}
	\centering
	\includegraphics[width=\textwidth]{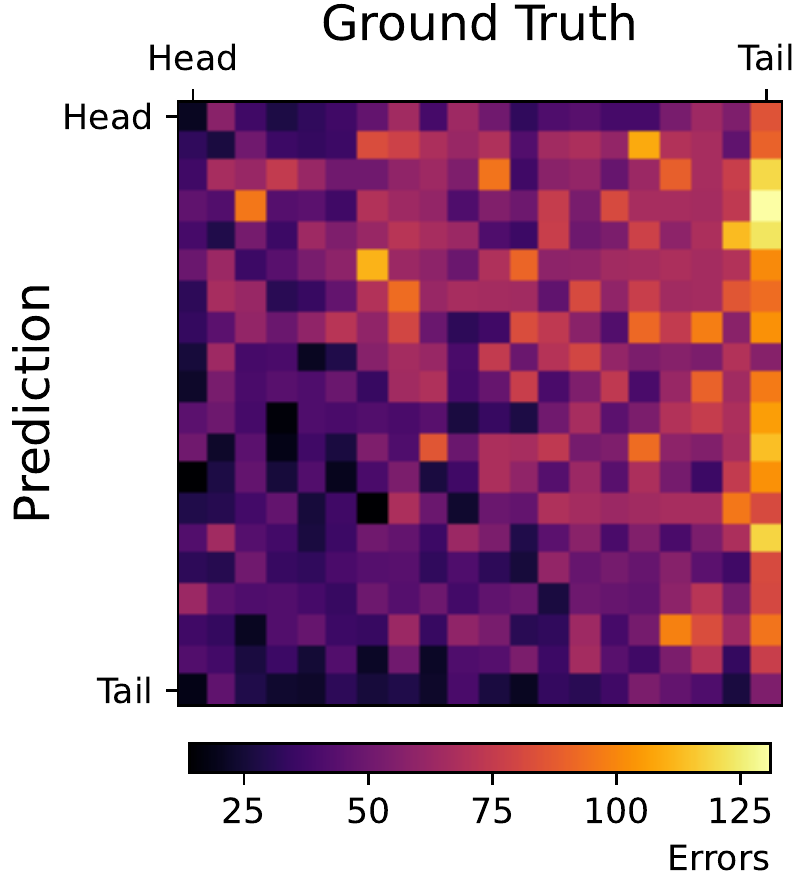}
	\caption{\small PaCo + BalSoftmax on ImageNet w/o DLSA}
	\label{fig:without}
\end{minipage}
\hfill
\begin{minipage}{0.3\textwidth}
	\centering
	\includegraphics[width=\textwidth]{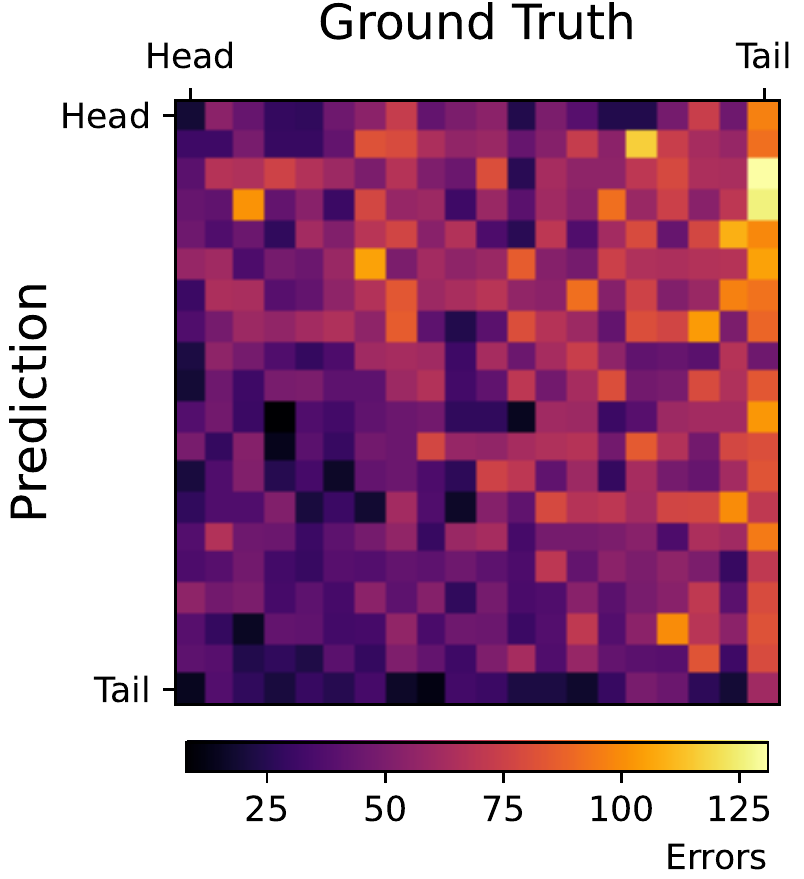}
	\caption{\small PaCo + BalSoftmax on ImageNet w/ DLSA}
	\label{fig:with}
\end{minipage}
\hfill
\begin{minipage}{0.3\textwidth}
    \centering
	\includegraphics[width=\textwidth]{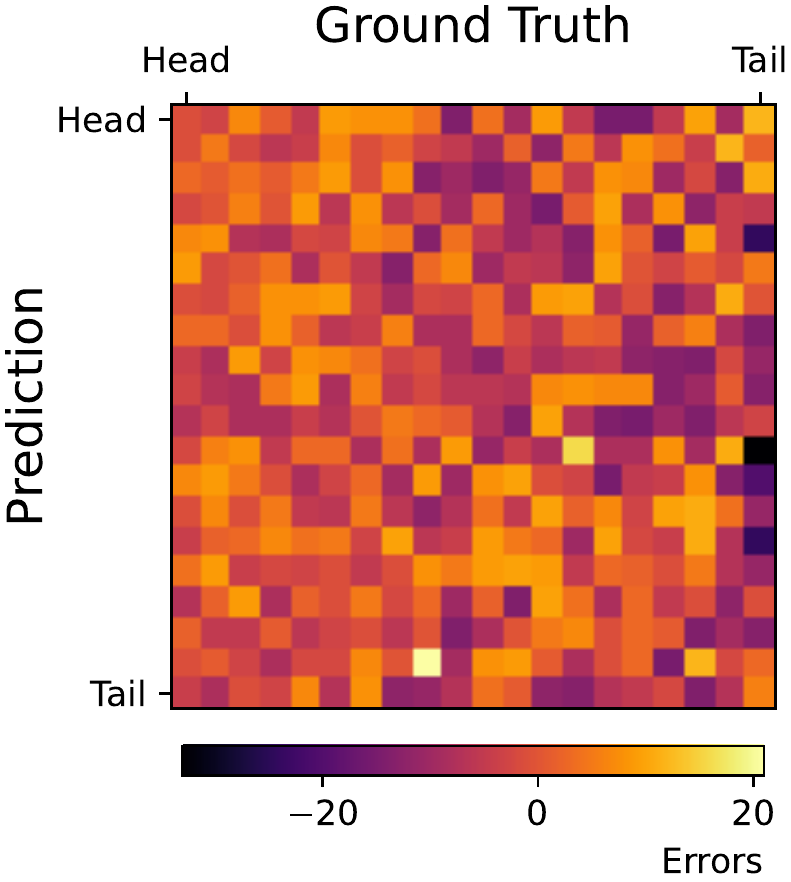}
	\caption{\small Difference of PaCo w/o and w/ DLSA}
	\label{fig:delta}
\end{minipage}
\hfill
\end{figure*}

The confusion matrix of PaCo+BalSoftmax w/ or w/o DLSA on ImageNet-LT are above. To illustrate the effect of DLSA, we also show the difference of Fig.~\ref{fig:with} and Fig.~\ref{fig:without} in Fig.~\ref{fig:delta}. The darker colors indicate the reduction of errors and improvement of accuracy. In Fig.~\ref{fig:delta}, the dark areas are mainly at the right top and middle since the DLSA reduces the error of misclassifying head classes to tail classes.

\section{Hyper-parameters}

We use cross-validation to select hyper-parameters. Different values are adopted on the datasets due to their different class number and granularity, unbalancedness and feature distribution.

Specifically, for ImageNet-LT, we use ResNet-50~\cite{resnet} as the backbone. The backbone is trained from scratch with the feature learning methods following the previous methods. 
Each Flow Filter has 500 clusters and is learned with learning rate 0.2 and batch size 1024 for 50 epochs. The loss weights are $\lambda_{bal}=1$, $\lambda_{pure}=0.02$.

For Places-LT, we use ImageNet~\cite{imagenet} pretrained ResNet-152 as the backbone. The Flow Filters are learned with learning rate 0.1 and batch size 512 for 60 epochs. The loss weights are $\lambda_{bal}=2$, $\lambda_{pure}=0.03$.

For iNaturalist18, we use ImageNet pretrained ResNet-50 as the backbone. The Flow Filters are learned with learning rate 0.2 and batch size 1024 for 30 epochs. The loss weights are $\lambda_{bal}=1$, $\lambda_{pure}=0.05$.

\section{Results on iNaturalist18}

The detailed results on iNaturalist18 (Many/Med/Few-show accuracy, MCC, NMI) are shown in Tab.~\ref{tab:inat-res}.

\begin{table}[h]
    \centering
    \caption{Results on iNaturalist18~\cite{van2018inaturalist} with ImageNet~\cite{imagenet}-pretrained ResNet-50~\cite{resnet}. }
    \adjustbox{width=0.75\linewidth}{
    \begin{tabular}{l|cccc|cc}
    	\toprule
    	Feature & Overall & Many & Medium & Few & MCC & NMI \\
    	\midrule
        LWS~\cite{decouple} & 69.5 & 71.0 & 69.8 & 68.8 & - & - \\
        cRT~\cite{decouple} & 68.2 & 73.2 & 68.8 & 66.1 & - & - \\
    	\midrule
    	PaCo+BalSoftmax~\cite{paco}        &  71.8  &  73.9  &  71.2  &  71.8  &  71.1  &  94.3 \\
    	PaCo+BalSoftmax~\cite{paco}+ DLSA  &  \textbf{72.8}  & \textbf{75.4}  &  \textbf{72.3}  &  \textbf{72.6}  &  \textbf{72.7} & \textbf{94.5} \\
    	\bottomrule
    \end{tabular}}
    \label{tab:inat-res}
\end{table}

\section{Ablation study on Places-LT}

\begin{table}[h]
    \centering
    \caption{Ablation study on Places-LT~\cite{openlongtail} with PaCo~\cite{paco}+BalSoftmax~\cite{BALMS} model and ResNet-152~\cite{resnet} backbone.}
    \adjustbox{width=0.5\linewidth}{
    \begin{tabular}{l|l|cccc}
    	\toprule
    	Method & Overall & Many & Medium & Few \\
    	\midrule
    	Full model      & \textbf{42.1} & \textbf{44.4} & \textbf{44.6} & \textbf{32.3} \\
    	\midrule
    	
    	w/o $\mathcal{L}_{\mathit{MLE}}$   & 40.8 & 43.7 & 43.2 & 30.1 \\
    	w/o $\mathcal{L}_{bal}$   & 41.2 & 44.1 & 43.7 & 30.2 \\
    	w/o $\mathcal{L}_{pure}$  & 41.5 & 44.0 & 43.9 & 31.2 \\
    	
    	\midrule
    	300 clusters    & 41.8 & 44.0 & \textbf{44.6} & 31.5 \\
    	1000 clusters   & 41.4 & 43.9 & 43.7 & 31.3 \\
    	
    	\bottomrule
    \end{tabular}}
    \label{tab:abl-place}
\end{table}

We extend the ablation study to Places-LT~\cite{openlongtail} on PaCo~\cite{paco}+BalSoftmax~\cite{BALMS} model. The results are shown in Tab.~\ref{tab:abl-place}.

\noindent{\bf Objectives.} Similar to ImageNet-LT, removing any loss ($\mathcal{L}_{\mathit{MLE}}$, $\mathcal{L}_{bal}$, $\mathcal{L}_{pure}$) leads to a significant performance drop. Among these losses, w/o $\mathcal{L}_{MLE}$ shows the greatest degradation since it controls the head-tail separation.

\noindent{\bf Cluster number.} Models with less/more clusters perform worse than default 500 clusters. Larger cluster number results in slow training and inference too.

\clearpage
\bibliographystyle{splncs04}
\bibliography{main}

\begin{thebibliography}{10}
\providecommand{\url}[1]{\texttt{#1}}
\providecommand{\urlprefix}{URL }
\providecommand{\doi}[1]{https://doi.org/#1}

\bibitem{buda2018systematic}
Buda, M., Maki, A., Mazurowski, M.A.: A systematic study of the class imbalance
  problem in convolutional neural networks. Neural Networks  \textbf{106},
  249--259 (2018)

\bibitem{byrd2019effect}
Byrd, J., Lipton, Z.: What is the effect of importance weighting in deep
  learning? In: International Conference on Machine Learning. pp. 872--881.
  PMLR (2019)

\bibitem{cai2021ace}
Cai, J., Wang, Y., Hwang, J.N.: Ace: Ally complementary experts for solving
  long-tailed recognition in one-shot. In: Proceedings of the IEEE/CVF
  International Conference on Computer Vision. pp. 112--121 (2021)

\bibitem{ACE}
Cai, J., Wang, Y., Hwang, J.N.: Ace: Ally complementary experts for solving
  long-tailed recognition in one-shot. In: Proceedings of the IEEE/CVF
  International Conference on Computer Vision. pp. 112--121 (2021)

\bibitem{LDAM}
Cao, K., Wei, C., Gaidon, A., Arechiga, N., Ma, T.: Learning imbalanced
  datasets with label-distribution-aware margin loss. arXiv preprint
  arXiv:1906.07413  (2019)

\bibitem{chao2015hico}
Chao, Y.W., Wang, Z., He, Y., Wang, J., Deng, J.: Hico: A benchmark for
  recognizing human-object interactions in images. In: Proceedings of the IEEE
  international conference on computer vision. pp. 1017--1025 (2015)

\bibitem{simclr}
Chen, T., Kornblith, S., Norouzi, M., Hinton, G.: A simple framework for
  contrastive learning of visual representations. In: International conference
  on machine learning. pp. 1597--1607. PMLR (2020)

\bibitem{Remix}
Chou, H.P., Chang, S.C., Pan, J.Y., Wei, W., Juan, D.C.: Remix: Rebalanced
  mixup. In: European Conference on Computer Vision. pp. 95--110. Springer
  (2020)

\bibitem{chu2020feature}
Chu, P., Bian, X., Liu, S., Ling, H.: Feature space augmentation for
  long-tailed data. In: Computer Vision--ECCV 2020: 16th European Conference,
  Glasgow, UK, August 23--28, 2020, Proceedings, Part XXIX 16. pp. 694--710.
  Springer (2020)

\bibitem{cubuk2020randaugment}
Cubuk, E.D., Zoph, B., Shlens, J., Le, Q.V.: Randaugment: Practical automated
  data augmentation with a reduced search space. In: Proceedings of the
  IEEE/CVF Conference on Computer Vision and Pattern Recognition Workshops. pp.
  702--703 (2020)

\bibitem{paco}
Cui, J., Zhong, Z., Liu, S., Yu, B., Jia, J.: Parametric contrastive learning.
  arXiv preprint arXiv:2107.12028  (2021)

\bibitem{danon2005comparing-nmi}
Danon, L., Diaz-Guilera, A., Duch, J., Arenas, A.: Comparing community
  structure identification. Journal of statistical mechanics: Theory and
  experiment  \textbf{2005}(09),  P09008 (2005)

\bibitem{imagenet}
Deng, J., Dong, W., Socher, R., Li, L.J., Li, K., Fei-Fei, L.: Imagenet: A
  large-scale hierarchical image database. In: 2009 IEEE conference on computer
  vision and pattern recognition. pp. 248--255. Ieee (2009)

\bibitem{RealNVP}
Dinh, L., Sohl-Dickstein, J., Bengio, S.: Density estimation using real nvp.
  arXiv preprint arXiv:1605.08803  (2016)

\bibitem{cosine2}
Gidaris, S., Komodakis, N.: Dynamic few-shot visual learning without
  forgetting. In: Proceedings of the IEEE Conference on Computer Vision and
  Pattern Recognition. pp. 4367--4375 (2018)

\bibitem{gupta2019lvis}
Gupta, A., Dollar, P., Girshick, R.: Lvis: A dataset for large vocabulary
  instance segmentation. In: Proceedings of the IEEE/CVF conference on computer
  vision and pattern recognition. pp. 5356--5364 (2019)

\bibitem{he2009learning}
He, H., Garcia, E.A.: Learning from imbalanced data. IEEE Transactions on
  knowledge and data engineering  \textbf{21}(9),  1263--1284 (2009)

\bibitem{moco}
He, K., Fan, H., Wu, Y., Xie, S., Girshick, R.: Momentum contrast for
  unsupervised visual representation learning. arXiv preprint arXiv:1911.05722
  (2019)

\bibitem{resnet}
He, K., Zhang, X., Ren, S., Sun, J.: Deep residual learning for image
  recognition. In: Proceedings of the IEEE conference on computer vision and
  pattern recognition. pp. 770--778 (2016)

\bibitem{DiVE}
He, Y.Y., Wu, J., Wei, X.S.: Distilling virtual examples for long-tailed
  recognition. arXiv preprint arXiv:2103.15042  (2021)

\bibitem{hong2021disentangling}
Hong, Y., Han, S., Choi, K., Seo, S., Kim, B., Chang, B.: Disentangling label
  distribution for long-tailed visual recognition. In: Proceedings of the
  IEEE/CVF Conference on Computer Vision and Pattern Recognition. pp.
  6626--6636 (2021)

\bibitem{huang2016learning}
Huang, C., Li, Y., Loy, C.C., Tang, X.: Learning deep representation for
  imbalanced classification. In: Proceedings of the IEEE conference on computer
  vision and pattern recognition. pp. 5375--5384 (2016)

\bibitem{huang2019deep}
Huang, C., Li, Y., Loy, C.C., Tang, X.: Deep imbalanced learning for face
  recognition and attribute prediction. IEEE transactions on pattern analysis
  and machine intelligence  \textbf{42}(11),  2781--2794 (2019)

\bibitem{FlowGMM}
Izmailov, P., Kirichenko, P., Finzi, M., Wilson, A.G.: Semi-supervised learning
  with normalizing flows. In: International Conference on Machine Learning. pp.
  4615--4630. PMLR (2020)

\bibitem{jamal2020rethinking}
Jamal, M.A., Brown, M., Yang, M.H., Wang, L., Gong, B.: Rethinking
  class-balanced methods for long-tailed visual recognition from a domain
  adaptation perspective. In: Proceedings of the IEEE/CVF Conference on
  Computer Vision and Pattern Recognition. pp. 7610--7619 (2020)

\bibitem{MetaDA}
Jamal, M.A., Brown, M., Yang, M.H., Wang, L., Gong, B.: Rethinking
  class-balanced methods for long-tailed visual recognition from a domain
  adaptation perspective. In: Proceedings of the IEEE/CVF Conference on
  Computer Vision and Pattern Recognition. pp. 7610--7619 (2020)

\bibitem{japkowicz2002class}
Japkowicz, N., Stephen, S.: The class imbalance problem: A systematic study.
  Intelligent data analysis  \textbf{6}(5),  429--449 (2002)

\bibitem{jiang2021self}
Jiang, Z., Chen, T., Mortazavi, B., Wang, Z.: Self-damaging contrastive
  learning. arXiv preprint arXiv:2106.02990  (2021)

\bibitem{kang2020exploring}
Kang, B., Li, Y., Xie, S., Yuan, Z., Feng, J.: Exploring balanced feature
  spaces for representation learning. In: International Conference on Learning
  Representations (2020)

\bibitem{decouple}
Kang, B., Xie, S., Rohrbach, M., Yan, Z., Gordo, A., Feng, J., Kalantidis, Y.:
  Decoupling representation and classifier for long-tailed recognition. arXiv
  preprint arXiv:1910.09217  (2019)

\bibitem{kiefer1952stochastic}
Kiefer, J., Wolfowitz, J.: Stochastic estimation of the maximum of a regression
  function. The Annals of Mathematical Statistics pp. 462--466 (1952)

\bibitem{kim2020adjusting}
Kim, B., Kim, J.: Adjusting decision boundary for class imbalanced learning.
  IEEE Access  \textbf{8},  81674--81685 (2020)

\bibitem{kim2020m2m}
Kim, J., Jeong, J., Shin, J.: M2m: Imbalanced classification via major-to-minor
  translation. In: Proceedings of the IEEE/CVF Conference on Computer Vision
  and Pattern Recognition. pp. 13896--13905 (2020)

\bibitem{IAF}
Kingma, D.P., Salimans, T., Jozefowicz, R., Chen, X., Sutskever, I., Welling,
  M.: Improved variational inference with inverse autoregressive flow. Advances
  in neural information processing systems  \textbf{29},  4743--4751 (2016)

\bibitem{alexnet}
Krizhevsky, A., Sutskever, I., Hinton, G.E.: Imagenet classification with deep
  convolutional neural networks. Advances in neural information processing
  systems  \textbf{25},  1097--1105 (2012)

\bibitem{MetaSAug}
Li, S., Gong, K., Liu, C.H., Wang, Y., Qiao, F., Cheng, X.: Metasaug: Meta
  semantic augmentation for long-tailed visual recognition. In: Proceedings of
  the IEEE/CVF Conference on Computer Vision and Pattern Recognition. pp.
  5212--5221 (2021)

\bibitem{li2022hake}
Li, Y.L., Liu, X., Wu, X., Li, Y., Qiu, Z., Xu, L., Xu, Y., Fang, H.S., Lu, C.:
  Hake: A knowledge engine foundation for human activity understanding. arXiv
  preprint arXiv:2202.06851  (2022)

\bibitem{li2020pastanet}
Li, Y.L., Xu, L., Liu, X., Huang, X., Xu, Y., Wang, S., Fang, H.S., Ma, Z.,
  Chen, M., Lu, C.: Pastanet: Toward human activity knowledge engine. In: CVPR
  (2020)

\bibitem{li2020symmetry}
Li, Y.L., Xu, Y., Mao, X., Lu, C.: Symmetry and group in attribute-object
  compositions. In: CVPR (2020)

\bibitem{li2021tlearning}
Li, Y.L., Xu, Y., Xu, X., Mao, X., Lu, C.: Learning single/multi-attribute of
  object with symmetry and group. TPAMI  (2021)

\bibitem{li2019transferable}
Li, Y.L., Zhou, S., Huang, X., Xu, L., Ma, Z., Fang, H.S., Wang, Y., Lu, C.:
  Transferable interactiveness knowledge for human-object interaction
  detection. In: CVPR (2019)

\bibitem{openlongtail}
Liu, Z., Miao, Z., Zhan, X., Wang, J., Gong, B., Yu, S.X.: Large-scale
  long-tailed recognition in an open world. In: Proceedings of the IEEE/CVF
  Conference on Computer Vision and Pattern Recognition. pp. 2537--2546 (2019)

\bibitem{tsne}
Van~der Maaten, L., Hinton, G.: Visualizing data using t-sne. Journal of
  machine learning research  \textbf{9}(11) (2008)

\bibitem{matthews1975comparison-mcc}
Matthews, B.W.: Comparison of the predicted and observed secondary structure of
  t4 phage lysozyme. Biochimica et Biophysica Acta (BBA)-Protein Structure
  \textbf{405}(2),  442--451 (1975)

\bibitem{menon2020long}
Menon, A.K., Jayasumana, S., Rawat, A.S., Jain, H., Veit, A., Kumar, S.:
  Long-tail learning via logit adjustment. arXiv preprint arXiv:2007.07314
  (2020)

\bibitem{info-nce}
Oord, A.v.d., Li, Y., Vinyals, O.: Representation learning with contrastive
  predictive coding. arXiv preprint arXiv:1807.03748  (2018)

\bibitem{MAF}
Papamakarios, G., Pavlakou, T., Murray, I.: Masked autoregressive flow for
  density estimation. arXiv preprint arXiv:1705.07057  (2017)

\bibitem{cosine1}
Qi, H., Brown, M., Lowe, D.G.: Low-shot learning with imprinted weights. In:
  Proceedings of the IEEE conference on computer vision and pattern
  recognition. pp. 5822--5830 (2018)

\bibitem{CLIP}
Radford, A., Kim, J.W., Hallacy, C., Ramesh, A., Goh, G., Agarwal, S., Sastry,
  G., Askell, A., Mishkin, P., Clark, J., et~al.: Learning transferable visual
  models from natural language supervision. In: International Conference on
  Machine Learning. pp. 8748--8763. PMLR (2021)

\bibitem{BALMS}
Ren, J., Yu, C., Sheng, S., Ma, X., Zhao, H., Yi, S., Li, H.: Balanced
  meta-softmax for long-tailed visual recognition. arXiv preprint
  arXiv:2007.10740  (2020)

\bibitem{normflow}
Rezende, D., Mohamed, S.: Variational inference with normalizing flows. In:
  International conference on machine learning. pp. 1530--1538. PMLR (2015)

\bibitem{samuel2021generalized-attribute}
Samuel, D., Atzmon, Y., Chechik, G.: From generalized zero-shot learning to
  long-tail with class descriptors. In: Proceedings of the IEEE/CVF Winter
  Conference on Applications of Computer Vision. pp. 286--295 (2021)

\bibitem{DRAGON}
Samuel, D., Atzmon, Y., Chechik, G.: From generalized zero-shot learning to
  long-tail with class descriptors. In: Proceedings of the IEEE/CVF Winter
  Conference on Applications of Computer Vision. pp. 286--295 (2021)

\bibitem{DRO-LT}
Samuel, D., Chechik, G.: Distributional robustness loss for long-tail learning.
  arXiv preprint arXiv:2104.03066  (2021)

\bibitem{samuel2021distributional}
Samuel, D., Chechik, G.: Distributional robustness loss for long-tail learning.
  arXiv preprint arXiv:2104.03066  (2021)

\bibitem{CBSampler}
Shen, L., Lin, Z., Huang, Q.: Relay backpropagation for effective learning of
  deep convolutional neural networks. In: European conference on computer
  vision. pp. 467--482. Springer (2016)

\bibitem{vgg}
Simonyan, K., Zisserman, A.: Very deep convolutional networks for large-scale
  image recognition. arXiv preprint arXiv:1409.1556  (2014)

\bibitem{sinha2020class}
Sinha, S., Ohashi, H., Nakamura, K.: Class-wise difficulty-balanced loss for
  solving class-imbalance. In: Proceedings of the Asian Conference on Computer
  Vision (2020)

\bibitem{spain2007measuring}
Spain, M., Perona, P.: Measuring and predicting importance of objects in our
  visual world  (2007)

\bibitem{van2018inaturalist}
Van~Horn, G., Mac~Aodha, O., Song, Y., Cui, Y., Sun, C., Shepard, A., Adam, H.,
  Perona, P., Belongie, S.: The inaturalist species classification and
  detection dataset. In: Proceedings of the IEEE conference on computer vision
  and pattern recognition. pp. 8769--8778 (2018)

\bibitem{wang2021rsg}
Wang, J., Lukasiewicz, T., Hu, X., Cai, J., Xu, Z.: Rsg: A simple but effective
  module for learning imbalanced datasets. In: Proceedings of the IEEE/CVF
  Conference on Computer Vision and Pattern Recognition. pp. 3784--3793 (2021)

\bibitem{wang2021contrastive}
Wang, P., Han, K., Wei, X.S., Zhang, L., Wang, L.: Contrastive learning based
  hybrid networks for long-tailed image classification. In: Proceedings of the
  IEEE/CVF Conference on Computer Vision and Pattern Recognition. pp. 943--952
  (2021)

\bibitem{wang2020RIDE}
Wang, X., Lian, L., Miao, Z., Liu, Z., Yu, S.X.: Long-tailed recognition by
  routing diverse distribution-aware experts. arXiv preprint arXiv:2010.01809
  (2020)

\bibitem{wang2017learning}
Wang, Y.X., Ramanan, D., Hebert, M.: Learning to model the tail. In:
  Proceedings of the 31st International Conference on Neural Information
  Processing Systems. pp. 7032--7042 (2017)

\bibitem{wu2020solving-taxonomy}
Wu, T.Y., Morgado, P., Wang, P., Ho, C.H., Vasconcelos, N.: Solving long-tailed
  recognition with deep realistic taxonomic classifier. In: European Conference
  on Computer Vision. pp. 171--189. Springer (2020)

\bibitem{LFME}
Xiang, L., Ding, G., Han, J.: Learning from multiple experts: Self-paced
  knowledge distillation for long-tailed classification. In: European
  Conference on Computer Vision. pp. 247--263. Springer (2020)

\bibitem{yang2020rethinking}
Yang, Y., Xu, Z.: Rethinking the value of labels for improving class-imbalanced
  learning. arXiv preprint arXiv:2006.07529  (2020)

\bibitem{zhang2021balanced}
Zhang, S., Chen, C., Hu, X., Peng, S.: Balanced knowledge distillation for
  long-tailed learning. arXiv preprint arXiv:2104.10510  (2021)

\bibitem{DisAlign}
Zhang, S., Li, Z., Yan, S., He, X., Sun, J.: Distribution alignment: A unified
  framework for long-tail visual recognition. In: Proceedings of the IEEE/CVF
  Conference on Computer Vision and Pattern Recognition. pp. 2361--2370 (2021)

\bibitem{FSR}
Zhang, Z., Pfister, T.: Learning fast sample re-weighting without reward data.
  In: Proceedings of the IEEE/CVF International Conference on Computer Vision.
  pp. 725--734 (2021)

\bibitem{MiSLAS}
Zhong, Z., Cui, J., Liu, S., Jia, J.: Improving calibration for long-tailed
  recognition. In: Proceedings of the IEEE/CVF Conference on Computer Vision
  and Pattern Recognition. pp. 16489--16498 (2021)

\bibitem{zhong2021improving}
Zhong, Z., Cui, J., Liu, S., Jia, J.: Improving calibration for long-tailed
  recognition. In: Proceedings of the IEEE/CVF Conference on Computer Vision
  and Pattern Recognition. pp. 16489--16498 (2021)

\bibitem{zhou2017places}
Zhou, B., Lapedriza, A., Khosla, A., Oliva, A., Torralba, A.: Places: A 10
  million image database for scene recognition. IEEE transactions on pattern
  analysis and machine intelligence  \textbf{40}(6),  1452--1464 (2017)

\bibitem{zhou2020bbn}
Zhou, B., Cui, Q., Wei, X.S., Chen, Z.M.: Bbn: Bilateral-branch network with
  cumulative learning for long-tailed visual recognition. In: Proceedings of
  the IEEE/CVF Conference on Computer Vision and Pattern Recognition. pp.
  9719--9728 (2020)

\bibitem{zipf2013psycho}
Zipf, G.K.: The psycho-biology of language: An introduction to dynamic
  philology. Routledge (2013)

\end{thebibliography}

\end{document}